\pdfoutput=1

\documentclass[11pt]{article}
\usepackage[dvipsnames]{xcolor}

\usepackage[preprint]{acl}

\usepackage{times}
\usepackage{latexsym}

\usepackage[T1]{fontenc}

\usepackage[utf8]{inputenc}

\usepackage{microtype}

\usepackage{inconsolata}

\usepackage{graphicx}
\usepackage{colortbl}
\usepackage{adjustbox}
\usepackage{textcomp}

\newcommand{\speeches}[1]{Speeches}
\newcommand{\novel}[1]{Novels}
\newcommand{\amt}[1]{AMT}
\newcommand{\blog}[1]{Blog}

\newcommand{\ourmethod}{\textsc{StyleRemix}\xspace}
\newcommand{\methodname}{\textsc{StyleRemix}\xspace}

\newcommand{\ourtestdataset}{\textsc{AuthorMix}\xspace}

\newcommand{\ourstyledatasetlong}{\textsc{Distilled Style Components Dataset}\xspace}
\newcommand{\ourstyledataset}{\textsc{DiSC}\xspace} %
\newcommand{\ourdataset}{\textsc{AuthorMix}\xspace}

\newcommand*\inlinelargeimage[1]{\raisebox{-0.3\baselineskip}{$\,$\includegraphics[height=1.25\baselineskip]{#1}$\,\,$}}
\newcommand{\largeemoji}{\inlinelargeimage{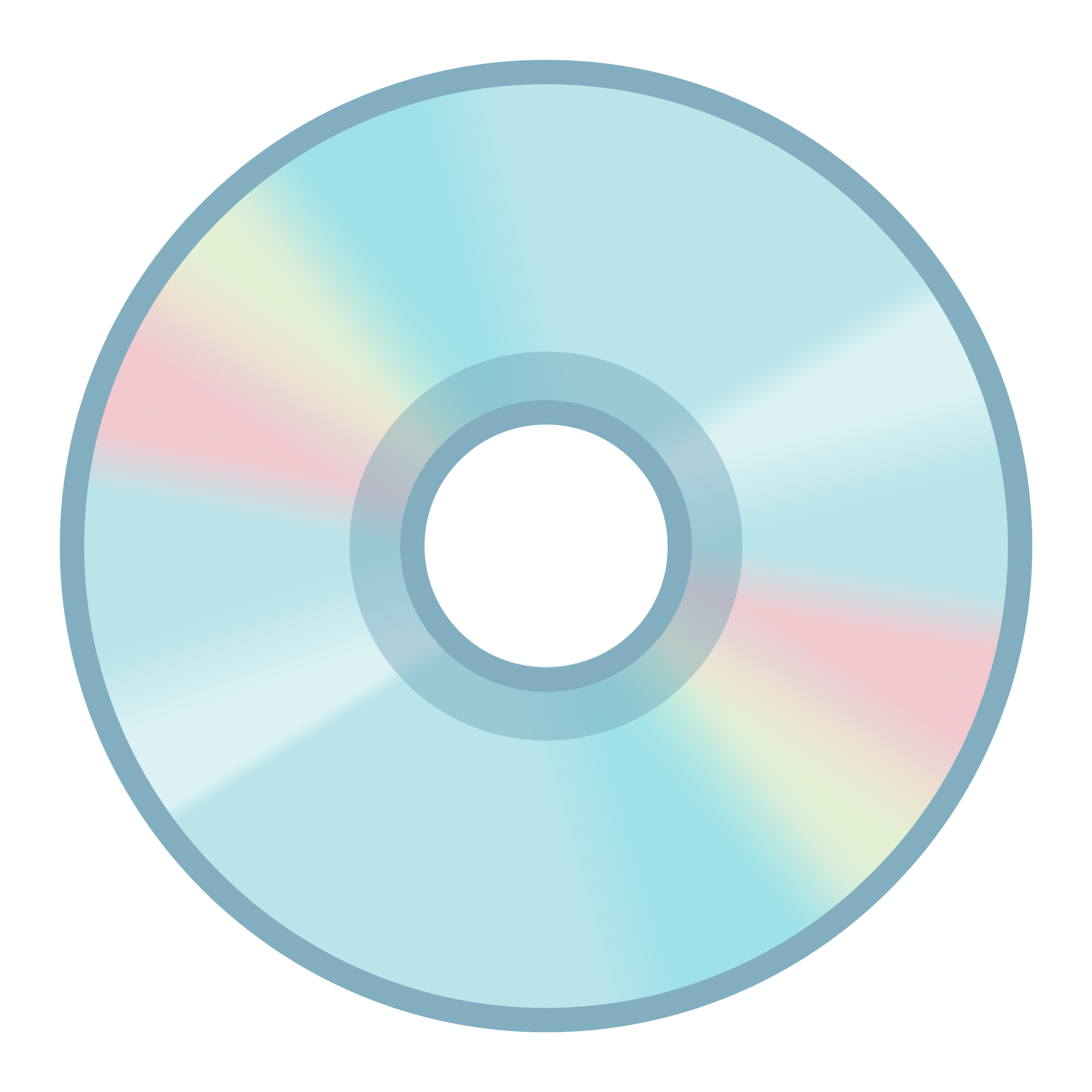}}

\newcommand*\inlinemediumimage[1]{\raisebox{-0.2\baselineskip}{$\,$\includegraphics[height=\baselineskip]{#1}$\,\,$}}
\newcommand{\mediumemoji}{\inlinemediumimage{Images/google-disk.png}}

\usepackage[T1]{fontenc}
\usepackage[utf8]{inputenc}
\usepackage{latexsym}
\usepackage{amsmath}
\usepackage{amssymb}
\usepackage{bm}
\usepackage{bbm}
\usepackage{graphicx}
\usepackage{breakcites}
\usepackage{amsthm}
\usepackage{mdframed}
\usepackage{lipsum}
\usepackage{algorithm}  
\usepackage[noend]{algpseudocode}
\usepackage[english]{babel}
\usepackage{delimset} %
\usepackage{tikz}

\usepackage{url}            %
\usepackage{booktabs}       %
\usepackage{amsfonts}       %
\usepackage{microtype}      %
\usepackage{wrapfig}
\usepackage{caption}

\usepackage{minitoc} %
\usepackage{xspace}
\usepackage{enumitem}
\usepackage{cleveref}

\usepackage{amsthm}
\usepackage{graphicx}
\usepackage{latexsym}
\usepackage{mathtools}
\usepackage{multirow}
\usepackage{fancyvrb}
\usepackage{adjustbox}

\crefname{section}{Section}{Sections}
\crefname{appendix}{Appendix}{Appendices}

\crefname{theorem}{Theorem}{Theorems}
\crefname{lemma}{Lemma}{Lemmas}
\crefname{corollary}{Corollary}{Corollaries}
\crefname{proposition}{Proposition}{Propositions}
\crefname{definition}{Definition}{Definitions}
\crefname{assumption}{Assumption}{Assumptions}

\Crefname{algorithm}{Algorithm}{Algorithms}
\crefname{figure}{Figure}{Figures}
\crefname{table}{Table}{Tables}

\usepackage{url}            %
\usepackage{booktabs}       %
\usepackage{amsfonts}       %
\usepackage{nicefrac}       %
\usepackage{microtype}      %
\usepackage{xcolor}         %

\usepackage{amsmath}
\usepackage{amssymb}
\usepackage{xspace}
\usepackage{enumitem}
\usepackage{bbm}
\usepackage{lipsum}  
\usepackage{multirow}
\usepackage{booktabs}
\usepackage{eqparbox}
\usepackage{scalerel}
\usepackage{float}
\usepackage{wrapfig}
\usepackage{caption}
\usepackage{comment}
\usepackage{subcaption}
\usepackage{tabularx}
\usepackage{algorithm}
\usepackage{algpseudocode}
\usepackage{ragged2e}

\definecolor{platinum}{rgb}{0.9, 0.89, 0.89}

\usepackage{amssymb}%
\usepackage{pifont}%

\usepackage{longtable}
 
\title{
\largeemoji StyleRemix: Interpretable Authorship Obfuscation via \\
Distillation and Perturbation of Style Elements
}
\newcommand{\aspace}{\hspace{1em}}
\newcommand{\uw}{$^{\heartsuit}$}
\newcommand{\aitwo}{$^{\clubsuit}$}

\author{Jillian Fisher\thanks{Co-first authors}\uw \aspace
  Skyler Hallinan\footnotemark[1]\uw \aspace 
  Ximing Lu\uw \aitwo \aspace
    Mitchell Gordon\uw \aspace  \\
      \textbf{Zaid Harchaoui}\uw \aspace
      \textbf{Yejin Choi}\uw \aitwo \aspace
      \\
      \uw University of Washington \aspace \aitwo Allen Institute for AI \\
      \texttt{\{jrfish, hallisky\}@uw.edu}
  }

\begin{document}
\maketitle

\begin{abstract}
Authorship obfuscation, rewriting a text to intentionally obscure the identity of the author, is an important but challenging task. 
Current methods using large language models (LLMs) lack interpretability and controllability, often ignoring author-specific stylistic features, resulting in less robust performance overall.

To address this, we develop \ourmethod, an adaptive and interpretable obfuscation method that perturbs specific, fine-grained \emph{style elements} of the original input text. \methodname uses pre-trained Low Rank Adaptation (LoRA) modules to rewrite an input specifically along various stylistic axes (e.g., formality and length) while maintaining low computational cost. \ourmethod outperforms state-of-the-art baselines and much larger LLMs in a variety of domains as assessed by both automatic and human evaluation. 

Additionally, we release \ourdataset, a large set of 30K high-quality, long-form texts from a diverse set of 14 authors and 4 domains, and \ourstyledataset, a parallel corpus of 1,500 texts spanning seven style axes in 16 unique directions\footnote{We release 1) our code at \url{https://github.com/jfisher52/StyleRemix} 2) a demo of \ourmethod at \url{https://huggingface.co/spaces/hallisky/StyleRemix} and 3) the datasets (\ourdataset and \ourstyledataset) and trained models in a  \href{https://huggingface.co/collections/hallisky/authorship-obfuscation-66564c1c1d59bb62eaaf954f}{HuggingFace collection}}.
\end{abstract}

\begin{figure}[h!]
    \centering
    \includegraphics[trim=0 0 0 16, clip, width = 0.97\linewidth]{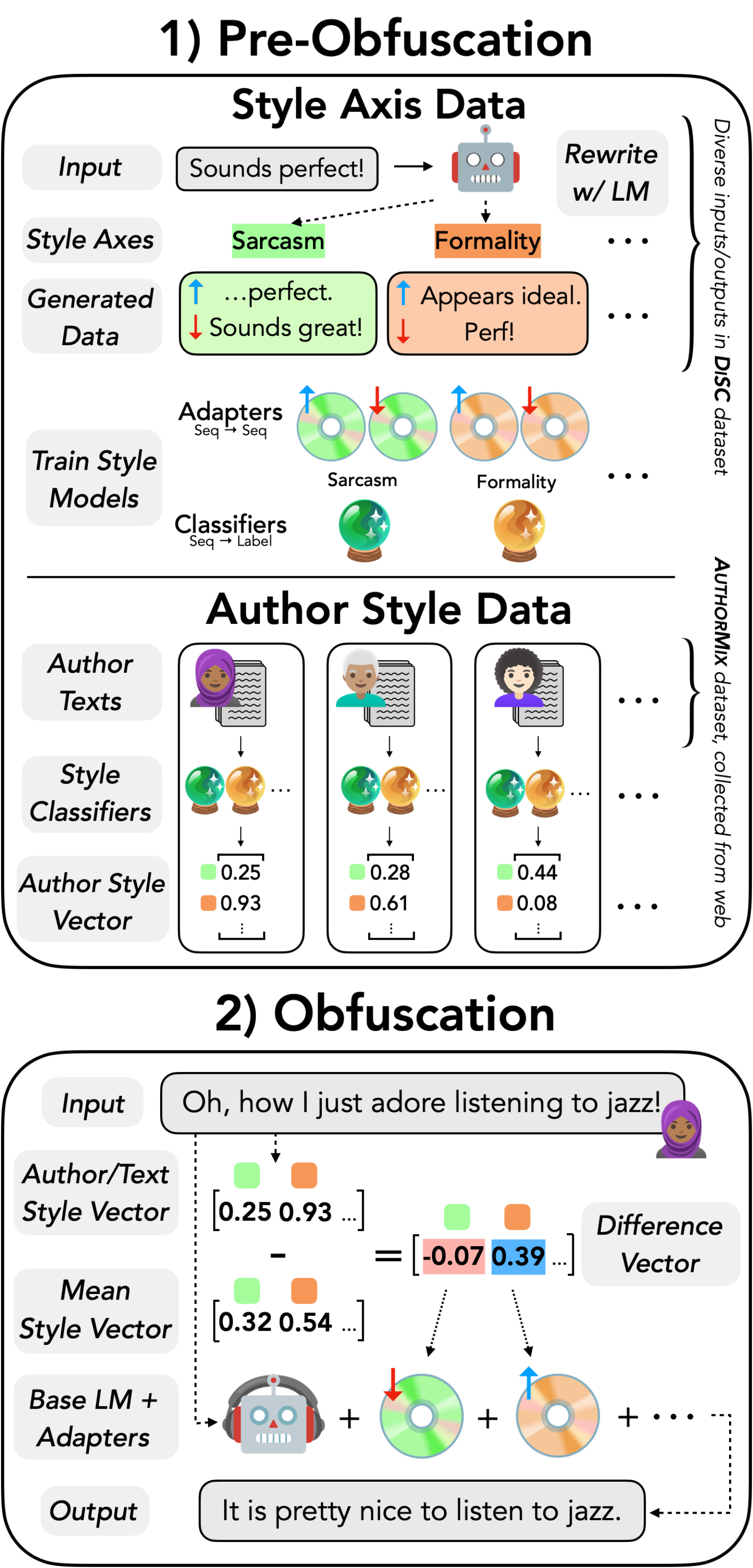}
    \caption{Overview of \mediumemoji \ourmethod. In pre-obfuscation, distinct \emph{style elements} are distilled from an LM into individual training sets, which are used to train specialized LoRA adapters. During obfuscation, the user can automatically or manually select the style adapter(s) which, when combined with the base LM, will best steer generations away from the original style.}
    \label{fig:method_overview}
\end{figure} %
\section{Introduction}

Authorship obfuscation, the act of rewriting text to conceal the author, is an important method
for preserving the privacy of authors in sensitive contexts, e.g., anonymous discussion forums, double-blind reviews, and health services.
However, it is inherently complex, requiring a substantial change in writing style to obscure the author's identity while also preserving the original content and fluency.

Historically, authorship obfuscation methods have manipulated aspects of an author's style to obfuscate the original text \cite{stylo_method, Shetty2017A4NTAA, Bevendorff2019HeuristicAO}. These techniques typically use style aspects that are easy to automatically evaluate, such as text length, capitalization frequency, and punctuation, to alter the original text. However, these rule-based methods are often too rigid and lead to a degradation of fluency and grammaticality \cite{fisher2024jamdec}.

\begin{figure*}[t!]
    \centering
    \includegraphics[width = \linewidth]{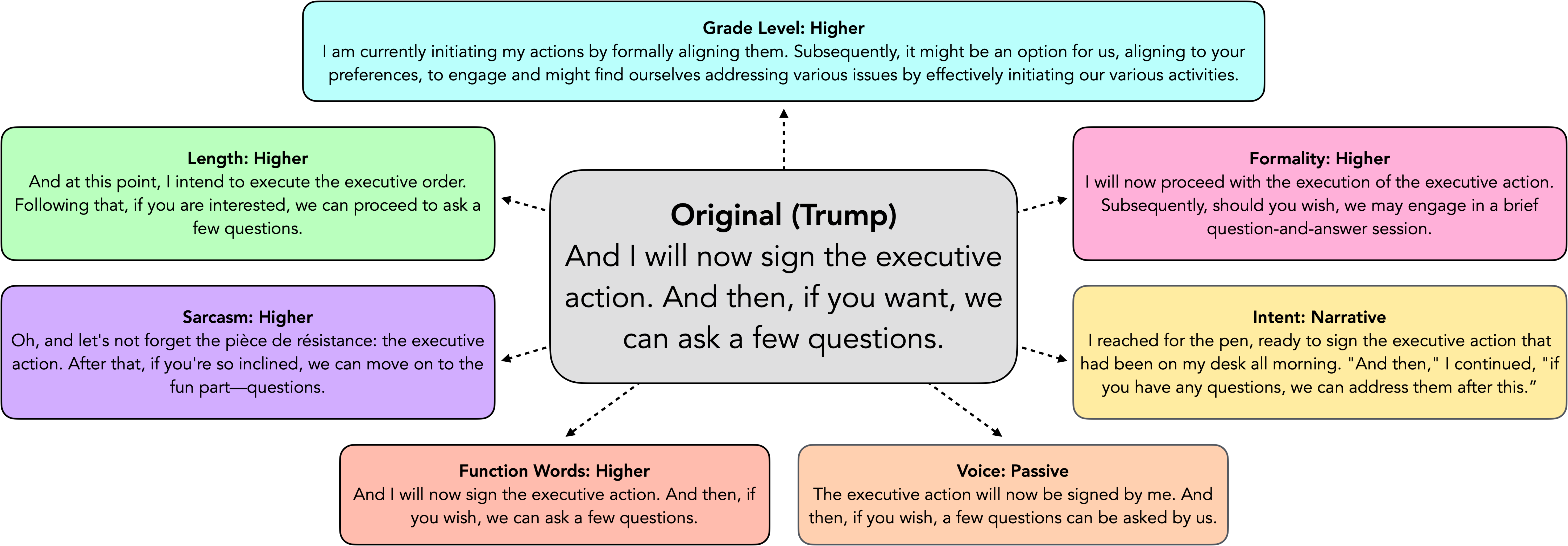}
    \caption{We compare generations from rewriting a text from \ourtestdataset-Speech using each of the style axis adapters \textit{individually}. This demonstrates the distinct transformation capabilities of each adapter, highlighting variations in tone, formality, and other linguistic features. We choose the direction of the style axes based on the automatic style selection method described in \cref{sect:stage_2_obf}.}
    \label{fig:qual_steerability}
\end{figure*}

Recent work demonstrates strong obfuscation performance using LLMs \cite{Mahmood2019AGH, avengers, weggenmann, fisher2024jamdec}, but the common challenge among these is a relative lack of interpretability and controllability on the obfuscation;
these approaches do not incorporate any author-specific stylometric characteristics 
of the original author, leading to more generalized and ineffective obfuscations.
For example, a method that relies solely on increasing language model fluency might effectively obfuscate more informal writing, but not formal writing.

To address this gap, we introduce \mediumemoji \ourmethod, an interpretable, inference-time, author-specific obfuscation method that combines the fluency and steerability of LLMs with author-specific style information. \ourmethod first detects unique stylistic elements of the author, either through automatic processes or manually. It then uses this information during obfuscation by integrating style-specific adapters with a base language model (an LLM) to guide the generated text away from the author's original style.

\ourmethod avoids high computational costs by utilizing pre-trained Low Rank Adaptation modules (LoRA; \citealp{Hu2021LoRALA}), which we train to rewrite inputs towards specific directions on various stylistic axes (e.g., more/less length, more/less formality, higher/lower grade level). 
Drawing inspiration from the process of creating a \emph{remix}, where musical elements of a song, such as tempo, key, and instrumentation are adjusted to form an entirely new track, in this work we seek to identify and manipulate different elements of an authorship style, and propose a simple yet effective approach to steer different components of the text with LoRA adapters. 
Our results show that \ourmethod outperforms state-of-the-art authorship obfuscation methods and instruction-based models of similar and larger sizes. Additionally, our method has the added benefit of explainability and is customizable to any unique authorship style.

We make the following contributions:
\begin{enumerate}[label=(\Roman*),nosep]
    \item We introduce \ourmethod, an interpretable, inference-time algorithm designed for authorship obfuscation. This method offers the personalization and flexibility required for effectiveness across various styles and text types.

    \item We release two datasets:\\ (1) \ourtestdataset, a comprehensive authorship dataset with over 30K paragraphs spanning four diverse domains (presidential speeches, novels, scholarly articles, and blogs) and 14 author styles, encompassing many more domains and styles than any previous work to our knowledge.
    \\(2) \ourstyledatasetlong (\ourstyledataset), a high-quality, validated, parallel dataset over 7 style axes. It features $n =$ 1,500 texts rewritten towards 16 distinct style directions for a total corpus size of 24K. %

\end{enumerate}
\section{Methods}

\ourmethod is an obfuscation method that leverages \emph{style elements} to adapatively rewrite texts. Specifically, it incorporates %
information about the style of the original author to guide the obfuscation process. Figure \ref{fig:method_overview} illustrates this new approach, which consists of two phases. 

The pre-obfuscation phase, conducted only once regardless of the number of authors, involves creating a diverse training set for each style axis we aim to modify (e.g., length variations, formality levels, grade level adjustments, etc.). These style-specific datasets are then used to train Low-Rank Adaptation (LoRA) adapters, which are low-parameter modules that can be seamlessly integrated with a larger base model to guide text generation along specific style axes. 

In the obfuscation phase, users can choose the style axes that most effectively disguise the original author's style, either automatically or manually. The selected pre-trained LoRA adapters are then used to steer the obfuscated text generation.

\subsection{Stage 1: Pre-Obfuscation}
\label{preobf}

\paragraph{Style Axes} When selecting the style axis, our goal is to identify ``author invariants'', which are text properties that are unique to a specific author. The widely accepted author invariants in the field of stylometry (the study of authorship style) include text length and the use of function words\footnote{Function words are words that express grammatical relationships among other words (if, up, would, etc.).} \cite{function_words}. Additionally, we incorporate "grade level," which primarily measures discrete features like the number of syllables and sentence and word lengths. Since this measure can vary slightly, we averaged three similar metrics: the Flesch-Kincaid (FK; \citealp{flesch1948new}), Linsear Write (L; \citealp{linsear}), and the Gunning Fog Index (GF; \citealp{gunning1952technique}) metrics. For the exact formulas, see \cref{appx:style_data_eval}.

Beyond formula-based properties, we also explore more abstract style axes such as the use of sarcasm, formality, voice (passive or active), and writing type
 (persuasive, descriptive, narrative, and expository). Due to the lack of existing formulas, we train model-based classifiers to measure these properties. More details on the training of these models can be found in \cref{appx:style_data_eval}.

In total, we identify seven style axes, each with two directions ("higher" or "lower"), except writing style, which has four options. This results in 16 distinct style elements. We acknowledge that this is not an exhaustive list of all author invariants, but we observed noticeable differentiation among the authors in our experimentation using these metrics. For more details, see \cref{appx:author_style_vector_analysis}.

\paragraph{Adapter Training Data} With the above style axes chosen, we create \ourstyledataset, a 16-style-element parallel dataset which distills each style element from a large LLM. To standardize the style adapter and minimize content dependencies, we create a single base training set and used instruction prompting with a LLM to generate rewrites along the chosen style axes. The base dataset comprises a diverse range of domains to encompass different writing types. Specifically, we randomly sample 500 paragraphs from sources including Wikipedia, books and plays \cite{kryscinski2021booksum}, and diary-style blogs \cite{schler2006effects}. Each paragraph is cleaned and standardized, resulting in paragraphs of 2-5 sentences each. Using GPT-4 Turbo \cite{openai2023gpt4turbo}, we then generate new versions of these paragraphs along different style axes and directions ("higher" or "lower") using detailed instruction prompt tuning (see \cref{appx:training_data_details}). This results in 16 parallel datasets written in different style axis and directions.

Next, we evaluate the generated paragraphs to ensure that they accurately reflect the intended style axis and direction. \cref{tab:train_data_eval} presents the evaluation results, both automatic and human, for the style training datasets created. The results demonstrate that our datasets effectively capture the desired styles. See \cref{appx:training_data_details} for more details.
\begin{table}[t!]
    \centering
    \begin{adjustbox}{width=\linewidth}
    \begin{tabular}{llll}
    \toprule
        \textbf{Style Axis} \textit{(metric)} & \textbf{Orig.} & \textbf{More} & \textbf{Less} \\ \midrule
        \textbf{Length} (\textit{words/sent}) & 18.87 & \textbf{23.04} & \underline{18.24} \\ 
        \textbf{Function Words} (\textit{\# func. words}) & 40.08 & \textbf{55.19 }& \underline{21.47} \\ 
        \textbf{Grade Level} \textit{(avg. FK, L, GF)} & 9.45 & \textbf{11.08} & \underline{6.72} \\ 
        \textbf{Formality} (\textit{model score}) & 0.68 & \textbf{0.97} & \underline{0.43} \\ \midrule
        ~ & \multicolumn{3}{c}{\textbf{Accuracy} \textit{(human eval)}} \\ \midrule
        \textbf{Sarcasm} & \multicolumn{3}{c}{97.7}  \\ 
        \textbf{Voice} & \multicolumn{3}{c}{93.7}  \\ 
       \textbf{Writing Intent} \textit{(4 classes)} & \multicolumn{3}{c}{77.7} \\ \bottomrule
    \end{tabular}
    \end{adjustbox}
    \caption{Evaluation of the parallel style training datasets. Automatic evaluation (top) is shown for the original score, as well as the score for the dataset that had instruction to increase (More) or decrease (Less) the given style axis. The \textbf{highest value} is bolded and the \underline{lowest value} is underlined. Other style axes required human evaluation (below). For this we randomly combine 10\% of the high and low datasets (or all four types for Writing Type) and ask three NLP experts to label whether the style axis was high or low; average accuracy is shown. }
\label{tab:train_data_eval}
\end{table}

\begin{table*}[t!]
\begin{adjustbox}{width=\textwidth}
\begin{tabular}{lcccccccccccc}
\toprule
\textbf{Model}     & \multicolumn{2}{c}{\textbf{Llama-2-Chat}}            & \multicolumn{2}{c}{\textbf{Llama-3-Inst}}          & \multicolumn{1}{l}{\textbf{Gemma-Inst}} & \multicolumn{1}{l}{\textbf{Paraphrase}} & \multicolumn{1}{l}{\textbf{MT}} & \multicolumn{1}{l}{\textbf{Stylo}} & \textbf{JD} & \multicolumn{3}{c}{\ourmethod} \\
\cmidrule(lr){2-3} \cmidrule(lr){4-5} \cmidrule(lr){6-6} \cmidrule(lr){11-13}
\textbf{Size}      & \multicolumn{1}{c}{\textbf{7B}}     & \multicolumn{1}{c}{\textbf{13B}} & \multicolumn{1}{c}{\textbf{8B}}     & \multicolumn{1}{c}{\textbf{70B}} & \multicolumn{1}{c}{\textbf{7B}}    & \multicolumn{1}{l}{}                     & \multicolumn{1}{l}{}            & \multicolumn{1}{l}{}               &             & \multicolumn{1}{l}{\textbf{Seq.}} & \multicolumn{1}{l}{\textbf{AM}} & \multicolumn{1}{l}{\textbf{AM + LoraHub$^+$}} \\
         \rowcolor{gray!25}
        \multicolumn{13}{c}{\ourdataset -- \textbf{Speech}} \\
\textbf{Drop Rate} & 18.2   & 24.0 & 17.6&16.8    & 23.1  & 24.1 & 10.3 & 15.1  &   29.2   & 34.9& 41.2 & 31.4     \\
\textbf{Grammar}   & 67.8   & 67.1 &67.1 &70.2    & 67.8  & 71.2 & 54.9 & 37.8  &   56.7   & 61.7& 66.5 & 63.9     \\
\textbf{Content}   & 83.8   & 80.8 &80.8 &80.2    & 78.6  & 83.9 & 89.1 & 89.5  &    56.4 & 71.3& 77.3 & 73.9     \\
\textbf{Overall}   & 10.3   & 13.0 & 9.5&9.5    & 12.3  & 14.4 & 5.1&  5.1   &  9.4   & \underline{15.3}  & \textbf{21.2} & 14.8     \\ \midrule
  \rowcolor{gray!25}
 \multicolumn{13}{c}{\ourdataset -- \textbf{Novels}} \\
\textbf{Drop Rate} & 12.2   & 13.7 &   9.2  & 11.3 & 13.3  & 10.8 & 7.0& 13.5  &    24.9  & 19.3& 28.6 & 35.6     \\
\textbf{Grammar}   & 71.8   & 73.8 &  73.1  & 75.4 & 70.0  & 68.3 & 46.3 & 36.8  &   61.2   & 72.6& 68.1 & 63.5     \\
\textbf{Content}   & 82.9   & 80.7 &   83.1 & 81.5 & 81.9  & 81.3 & 85.2 & 88.1  &    58.6  & 83.7& 76.1 & 72.9     \\
\textbf{Overall}   & 7.3    & 8.2&   5.6  & 6.9& 7.6   & 6.0  & 2.8& 4.4   &    8.9  & 11.8& \underline{14.8} & \textbf{16.5}     \\ \midrule
         \rowcolor{gray!25}
        \multicolumn{13}{c}{\ourdataset -- \textbf{Scholar}} \\
\textbf{Drop Rate} & 0.8    & 1.5& 1.6    & 2.5& 0.0   & 0.8  & 1.5& 4.6   &   6.1   & 1.8 & 9.2& 11.5     \\
\textbf{Grammar}   & 64.3   & 64.9 & 64.1   & 66.6 & 65.3  & 69.1 & 54.5 & 31.0  &   62.3   & 65.8& 48.6 & 44.7     \\
\textbf{Content}   & 91.7   & 89.7 & 88.9   & 84.0 & 88.9  & 91.3 & 92.8 & 85.8  &   60.6  & 78.0& 75.3 & 68.8     \\
\textbf{Overall}   & 0.5    & 0.9& 0.9    & 1.4& 0.0   & 0.5  & 0.8& 1.2   &  2.3  &      0.9  &\underline{ 3.4}& \textbf{3.5}      \\ \midrule
         \rowcolor{gray!25}
        \multicolumn{13}{c}{\ourdataset -- \textbf{Blog}} \\
\textbf{Drop Rate} & 17.7 & 21.3 & 21.8  & 18.9 & 27.5 & 22.2  & 9.4 & 12.1  &   56.4   & 34.4& 41.0 & 42.0     \\
\textbf{Grammar}   & 68.4   & 69.1 & 71.3   & 74.0 & 69.0  & 69.8 & 41.9 & 29.1  &   60.6  & 66.7& 64.9 & 65.3     \\
\textbf{Content}   & 82.5   & 79.0 & 78.1   & 77.8 & 77.8  & 80.4 & 83.7 & 85.8  &   45.1   & 72.1& 73.7 & 74.2     \\
\textbf{Overall}   & 10.0   & 11.6 & 12.1   & 10.9 & 14.8  & 12.5 & 3.3& 3.0   &   15.4   & 16.5& \underline{19.6} & \textbf{20.4}    \\ \bottomrule
\end{tabular}
\end{adjustbox}
    \caption{Comparison of obfuscation methods measured by mean drop rate, grammar, meaning similarity, and overall (the mean product of the metrics), across \ourmethod and comparatively sized or larger baselines on each subset of \ourdataset. \textbf{Bold} and \underline{underline} denote the highest and the second-highest score respectively in each row. All metrics displayed in the table are multiplied by 100 for easier viewing of significant figures.}
    \label{tab:auto_eval_baseline}
\end{table*} 
\paragraph{Train LoRA Adapters}
Next, our goal is to train the models to generate text along the chosen style axes.
To minimize computational cost \cite{Strubell2019EnergyAP}, we bypass model fine-tuning and instead employ \emph{Low Rank Adapation} (LoRA; \citealp{Hu2021LoRALA}) adapters for each of the style axes. By freezing the base model and tuning only a small portion of injected features, LoRA guarantees lightweight training \cite{Rebuffi2017LearningMV, Houlsby2019ParameterEfficientTL} while also incurring \textit{no additional inference latency}, ensuring both efficient training and deployment. We use Llama-3 8B \cite{llama3modelcard} as our base model, and train LoRA adapters on top of them for each direction on the style axes. See \cref{lora} for more training details. %

\subsection{Stage 2: Obfuscation}\label{sect:stage_2_obf}
\paragraph{Style Axes and Weights Selection}
During the obfuscation phase, a text or set of texts is presented for obfuscation. If a user has a clear idea of which style axes to adjust, they can input their desired styles and the corresponding weights of the adapters to control the strength of the generation. However, since this information is often unavailable, we develop a straightforward yet effective method for selecting which style axes to modify and the magnitude of the weights of these adapters.  %

For the given $m$ authors in some genre (e.g. speech, novel), we first create an author vector $\bm{x_i} \in \mathbb{R}^7$ for each author, which is composed of the automatic evaluation of the seven style axes. After normalizing with respect to all $m$ authors,
we calculate the ``difference'' vector between each author and the %
average, defined as $\bm{\bar{x}_i} = \bm{x_i} - \frac{1}{m} \sum_{j=1}^m \bm{x_j}$. Using the absolute values in this difference vector $\vert \bm{\bar{x}_i} \vert$, users can select the top $k$ style axes where the specific author deviates most from the average. 

Next, the user needs to specify the weight for each chosen style adapter to merge with the base model. This procedure could be manual, but we also provide a heuristic to determine the weights automatically. Building on prior work, we find that LoRA adapters perform well with values in the range [-1.5, 1.5] \cite{Huang2023LoraHubEC}. 
Next, we use the number of standard deviations an author vector deviates from the average to map each style axis to a set of predetermined weights $w_i$. Specifically,
\begin{align*}
\footnotesize
 w_i \begin{cases} 
      0.7 & \text{std}(\bar{x}_i)\leq 1\\
      0.9 & 1< \text{std}(\bar{x}_i) \leq 2 \\
      1.2 & 2< \text{std}(\bar{x}_i)\leq 3\\
      1.5 & \text{std}(\bar{x}_i) > 3
   \end{cases}
\end{align*}
For detailed implementation, see \cref{appx:our_method_style_weight_selection}.
\paragraph{Generation Techniques} During generation, we use the adapters corresponding to the selected style axes to rewrite the given text, steering these prominent styles toward the average. In addition, we experiment with multiple methods for combining these LoRA adapters.
\begin{itemize}[topsep=0pt, itemsep=0pt, partopsep=0pt,parsep=0pt]
\item \textbf{Sequential}: We pass in the text through a sequence of adapters iteratively; the output from one adapter serves as the input for the next. This method provides additional interpretability by revealing how the text becomes obfuscated at different stages after altering specific style axes. However, it increases computation time, as it requires a forward pass for each chosen style axis.

\item \textbf{Adapter Merging (AM)}: We merge the weights of all the adapters before combining them with the base model by \emph{concatenating} their weights \cite{yadav2023tiesmerging, yu2024language}. See \cref{concat} for more details.

\item \textbf{LoraHub$^{+}$}: LoraHub is a framework designed to assemble multiple LoRA adapters with the goal of maximizing performance on specific tasks \cite{Huang2023LoraHubEC}. It adjusts the weights of the given adapters to optimize the specified objective through gradient-free optimization. 
We extend this with \textbf{LoraHub$^+$}, which defines a new objective function $L$ designed to optimize for obfuscation by summing up the automatic evaluations of the selected style axes across a small set of test examples. We also add the fluency score to encourage more fluent text:
\begin{align*}
\footnotesize
 L &= \sum_{v_i \in \text{selected axes}}\begin{cases} 
      v_i & v_i \leq \frac{1}{m} \sum_{j=1}^m x_i\\
      1 - v_i & v_i > \frac{1}{m} \sum_{j=1}^m x_i
   \end{cases} \\ & \phantom{haha}+ \alpha \cdot s_f
\end{align*}

\vspace*{-1mm}
where $v_i$ represents the automatic evaluation for a selected style axis on the subset of test examples, $s_\text{f}$ represents fluency score, and $\alpha$ denotes the discount factor. LoraHub$^+$ is used \emph{in conjunction} with adapter merging.

\end{itemize} 
Merging multiple adapters with concatenation is  computationally efficient. Specifically, we find that merging four adapters with the base model (using AM) takes less than 5 seconds on average.

\section{Experiments}

\subsection{Datasets}
We aimed to test how authorship obfuscation methods perform on a diverse array of author styles and domains. To this end, we develop a new benchmark dataset called \ourdataset, covering four distinct domains: presidential speeches, early-1900s fiction novels, scholarly articles, and diary-style blogs. Together, \ourdataset contains more than $30$ k high-quality paragraphs from 14 authors. %

For the presidential domain, we curate and clean speeches from George W. Bush, Barack Obama, and Donald Trump\footnote{We select these presidents/authors due to their diverse styles but similar eras to minimize content discrepancies.}. For novel domain, we choose a collection of early 1900s fiction writers with strong writing styles: Ernest Hemingway, F. Scott Fitzgerald, and Virginia Woolf. We choose these specific writers in an effort to limit the topic bias in the evaluation metrics. 

Lastly, we alter two existing datasets to match the formality of our new domains: the Extended-Brennan Greenstad \cite{amt_dataset}, a collection of ``scholarly'' short (500-word) paragraphs gathered from Amazon Mechanical Turk (AMT), and the Blog Authorship corpus \cite{blog_dataset}, a collection of blogs (diary-style entries) that were posted to blog.com. %
More details can be found in \cref{appx:exp_data}. 

\subsection{\ourmethod Configurations} We compare three versions of \ourmethod: sequential, adapter merging, and LoraHub$^+$. For sequential, to account for the order of the styles, we average over $n=3$ shuffled orders.  The adapter merging method uses the static standard deviation mapping method described in \cref{sect:stage_2_obf}. For these two methods, we select the best method per domain (based on the overall score) using the top $k = 1,2,3,4$ changed styles. Lastly, we run our customized LoraHub method (LoraHub$^+$), matching the best styles per domain as the base adapter merging method for direct comparison. 

\subsection{Baselines}
We compare against both SOTA obfuscation methods and equal and larger-size LLMs using instructions. Full details can be found in \cref{appx:experimental_details}. 

\paragraph{Stylometric (Stylo)} We use the stylometric obfuscation technique presented by \citet{stylo_method}, which examines various statistical features that characterize a writer's style, such as sentence length and word frequency, and then modifies the text to align these features with an "average" value, which is established using a training dataset.

\paragraph{Machine Translation (MT)} \citet{Keswani2016AuthorMT} introduce \emph{round-trip machine translation} by translating a text from English to German, then to French, and then back to English. We use the new M2M translation models \cite{m2m100}.

\paragraph{Paraphraser (Paraphrase)} We use the T5-Large paraphraser introduced by \citet{jung2024impossible} which iteratively improves through self-distillation.

\paragraph{JAMDEC (JD)} This method \cite{fisher2024jamdec} relies on a smaller LLM, GPT2-XL \cite{gpt2} to overgenerate many new rewrites given the keywords from the original text. It then uses a filter to select the best new rewrite. We run this method using the default settings, and a beam width of $10$. 

\paragraph{Instruction-tuned LLMs} We compare against a suite of instruction-tuned LLMs including Llama-2-Chat (7B, 13B) \cite{llama2}, Llama-3-Instruct (8B, 70B) \cite{llama3modelcard}, and Gemma-Instruct (7B) \cite{gemmateam2024gemma}. For each model, we provide instruction to ``rewrite'' the given text. More comparisons of different models can be found in the \cref{appx:compare_more_llms}. Exact instructions used for generation can be found in \cref{appx:experimental_details}.

\subsection{Automatic Evaluations}
In line with previous work, we evaluate authorship obfuscation on four main criteria: obfuscation, content preservation, grammaticality, and overall\footnote{All metrics are bounded between 0 and 1, which ensures the product has the same bounds. Although similarity scores are theoretically bounded from -1 to 1, we observe empirically across all datasets and methods that they are bounded more strictly between 0 and 1; see \cref{appx:obf_classifier} for more details}. See \cref{appx:obf_classifier} for more details.

\paragraph{Obfuscation} Classifiers with various machine learning architectures have been used to measure obfuscation \cite{Mahmood2019AGH, avengers, fisher2024jamdec}. Recent work in authorship obfuscation and style transfer often uses RoBERTa \cite{Liu2019RoBERTaAR} classifiers \cite{xing2024alison, Uchendu2021TURINGBENCHAB, Liu2024StyleTW, Hallinan2023STEERUS}. 

In line with this previous work, we fine-tune four RoBERTa large (355M) models, one for each domain in \ourdataset; classifiers achieve on average $94.0\%$ accuracy in the evaluation set of each respective domain (for full results, see \cref{obf_appendix_classifier}).
Using these classifiers, we calculate the drop rate, which is the normalized decrease in the classifier's accuracy when comparing its performance on the original texts to the obfuscated texts. The drop rate can be expressed as: 
\begin{align*}
    \text{Drop Rate} = \frac{acc_{\text{orig}} - acc_{\text{obf}}}{acc_{\text{orig}}}
\end{align*}
We also report evaluation using an alternative metric to measure obfuscation based on universal authorship representations \cite{Soto2021LearningUA} in \cref{altmetrics}; these results
corroborate our main findings in \S \ref{mainresults}.

\paragraph{Content Preservation}
We use the embedding similarity of the inputs and their obfuscations in Sentence Transformers \cite{Reimers2019SentenceBERTSE} to gauge semantic similarity.

\paragraph{Language Quality} We measure grammaticality via the probability of being grammatically acceptable from TextAttack \cite{textattack}, a binary RoBERTa-large classifier fine-tuned on the Corpus of Linguistic Acceptability \cite{Warstadt2018NeuralNA}.

\paragraph{Overall Task Score} The overall success of each obfuscation is measured by the product of the above three metrics: drop rate, similarity score, and CoLA score. This product ensures a high overall task score accurately reflects high scores in all three categories; it is also used in prior work in text rewriting
\cite{Krishna2020ReformulatingUS, Hallinan2023STEERUS, patel2023lowresourceauthorshipstyletransfer, xu-etal-2018-unpaired} 

\subsection{Human Evaluation} We also conduct human evaluation to verify the quality of the obfuscations from the best \ourmethod variant and comparably sized baselines; we omit . We randomly select $n=20$ texts from each author in \ourdataset for annotation via Amazon Mechanical Turk by three workers each. Following the setup of \citet{fisher2024jamdec}, we instruct each annotator to read both the original and obfuscated text, then respond to five questions rated on a three-point Likert scale (0, 0.5, or 1), measuring grammar, fluency, high content preservation, low content addition, and obfuscation. We discard evaluations where all annotators disagree on the label\footnote{Pairwise agreement is greater than 93\% for all questions}. Lastly, we calculate an \emph{overall} score using the weighted product of all five metrics. Further details can be found in \cref{appx:appx:human_evals}. %
\begin{figure}[t!]
    \centering
    \includegraphics[width = \linewidth]{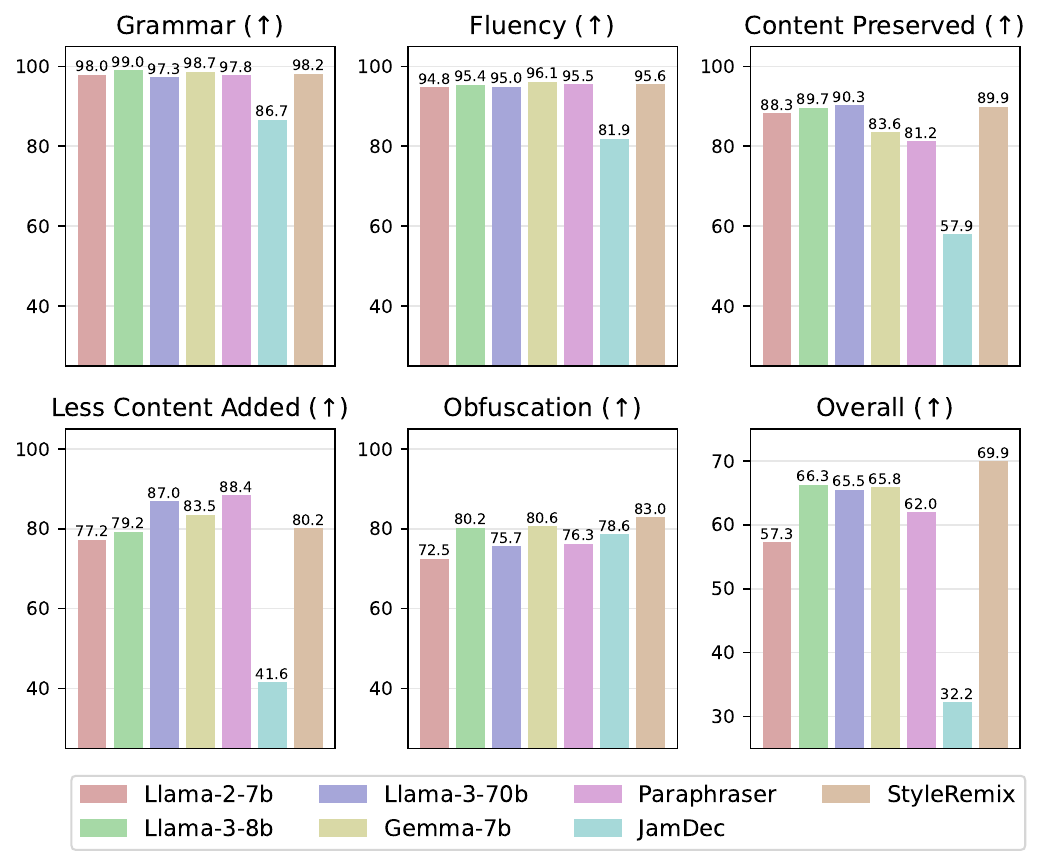}
    \caption{Human evaluation results for mean grammar, fluency, content preserved, less content added, and obfuscation. For each of the metrics, \emph{higher} is better. We also compute the mean overall score, the product of grammar, content preserved, and less style similarity.}
    \label{fig:human_eval}
\end{figure}
\subsection{Main Results}
\label{mainresults}

\paragraph{\ourmethod has the highest overall task score and obfuscation drop rate} \cref{tab:auto_eval_baseline} compares \ourmethod to LLMs of all sizes and the four SOTA authorship obfuscation methods on obfuscation, content preservation, grammar, and overall task score. \ourmethod has on average $5.5\%$ higher overall task score and up to $12\%$ higher obfuscation drop rate on all domains, with similar scores on content preservation and grammar. Impressively, \ourmethod is also more effective at obfuscation than LLMs of much larger magnitudes -- such as Llama-2-Chat 13B and Llama-3-Instruct 70B -- especially on the Blog domain with a $\sim\!10\%$ higher overall task score. This shows how a more targeted approach can outperform larger but more generic methods.
\begin{table}[t!]
\begin{adjustbox}{width=\linewidth}
\footnotesize
\centering
\begin{tabular}{p{1.29cm}p{9.4cm}}
\toprule
\textbf{Method} & \textbf{Generation} \\ \midrule
\rowcolor{gray!25} \textbf{Original (Blog 3)} &  100\%!  I was surprised, but not complaining lol.  But yeah Mr.  Curphey's now on crutches, hobbling around the classroom and still teaching us like his usual self. \\ \midrule
\textbf{MT} & \textcolor{red}{100 per cent!} I was surprised, but I don’t complain lol. But yes, Mr. Curphey is now in the \textcolor{red}{gardens to walk} through the classrooms and still teach us as his ordinary self. \\ 
\textbf{Stylo.} & \textcolor{red}{One hundred percent}; i was surprised, but not complaining ll, as but yeah \textcolor{red}{Mr. murphy} 's now on crutches, \textcolor{red}{sobbing} around the classroom and always teaching us like his usual self. \\ 
\textbf{JD} &  \textcolor{red}{100\%!  I was surprised, but not complaining lol.  But yeah Mr.  Curphey's now on crutches, hobbling around the classroom and still teaching us like his usual self.}\\
\textbf{Para.} & Mr. Curphey is currently using crutches in the classroom, but he is still teaching us with his usual enthusiasm \textcolor{red}{and enthusiasm}. I was taken aback, but I didn't feel \textcolor{red}{any remorse}. \\ \midrule
\textbf{Llama-2 Chat (7B)} & \textcolor{red}{100\%!} I was \textcolor{red}{pleasantly} surprised, but not complaining. However, Mr. Curphey is now using crutches to move around the classroom, still actively teaching us with his usual dedication. \\
\textbf{Llama-3 Inst (8B)} & \textcolor{red}{100\%!} I was taken aback, but I wasn't about to grumble about it, haha! But yes, Mr. Curphey's now on crutches, limping around the classroom, and yet he's still teaching us as himself, business as usual. \\
\textbf{Gemma Inst (7B) }& \textcolor{red}{100\%!} I was surprised, but not in a bad way. Mr. Curphey is still teaching us despite being on crutches. \\ 
\rowcolor{Green!25} \midrule
\textbf{\textsc{Style}} \textbf{\textsc{Remix}} & Initially, I experienced a \textcolor{Green}{notable degree of surprise}, yet my responses remain \textcolor{Green}{devoid of significant complaint}. Currently, Mr. Curphey, significantly, is navigating our classroom utilizing crutches, \textcolor{Green}{effectively maintaining} his usual style of instruction, significantly. \\ \midrule
\midrule \rowcolor{gray!25} \textbf{Original (Obama)} &  I recognize that times have changed since I first spoke to this convention. The times have changed, and so have I. I'm no longer just a candidate. I'm the President. \\ 
\textbf{MT} & I recognize that the times have changed since I first spoke to this convention. The times have changed, and \textcolor{red}{that’s how I have}. I’m not just a candidate anymore. I am the President. \\ 
\textbf{Stylo.} & I acknowledge that times have changed since I \textcolor{red}{1st} spoke to this convention, and the times have changed, and so have I. I am no longer just a candidate; and myself \textcolor{Red}{'m} the President. \\
\textbf{JD} & It's important to remember where we came from, where we're going, and what we're trying to do," he said, adding, ``I think it's important for all of us to be.  We are the people who built this country, and we have a responsibility to make sure that we preserve it for our children and our children's children, and that's what we're going to do,''. \\ 
\textbf{Para.}& Since I first addressed this convention, \textcolor{red}{I have seen significant changes. As a result}, I have evolved from a candidate to the current President. \\ \midrule
\textbf{Llama-2 Chat (7B)} & Since my initial conversation with this convention, time has passed and changes have occurred. As a result, \textcolor{red}{my perspective and circumstances have evolved}. I am no longer just a candidate, but the actual President. \\ 
\textbf{Llama-3 Inst (8B)} & Since my initial address to this esteemed convention, I have come to acknowledge that the times have indeed changed. So too have I. Gone are the days when I stood before you as a mere candidate, eager to make my case. Today, I stand before you as the President, \textcolor{red}{with a wealth of experience and insight gathered during my time in office}. \\ 
\textbf{Gemma Inst (7B)} & Since my initial address to this esteemed convention, the world has shifted, and I have evolved alongside it. Today, I stand before you not as a candidate, but as the President, \textcolor{red}{ready to tackle the challenges that lie ahead}. \\ \rowcolor{Green!25} \midrule
\textbf{\textsc{Style}} \textbf{\textsc{Remix}} & \textcolor{Green}{Oh, how the world has transformed} since I first addressed you all here! \textcolor{Green}{Indeed, the world has shifted}, and so have I. It’s not just about being a candidate anymore—\textcolor{Green}{it’s about being the President.} \\
\bottomrule
\end{tabular}
\end{adjustbox}
\caption{Examples of obfuscations from baseline methods and \ourmethod from texts from \ourdataset-Blog and \ourdataset-Speech respectively. \textcolor{Green}{Green} regions highlight where the method obfuscates well and \textcolor{red}{red} areas show grammar, content, or obfuscation issues.}
\label{tab:example_outputs}
\end{table} 
\paragraph{Human evaluation shows \ourmethod outputs the best overall and most obfuscated generations}

\cref{fig:human_eval} shows the result of our human evaluation. All methods have strong grammar and fluency scores, with only slight differences between them. \ourmethod beats all other baselines in obfuscation with a score of $83.0\%$, and ranks second for content preservation, trailing by just 0.4 points behind the significantly larger LLama-3-Instruct 70B. Notably, \ourmethod even surpasses the larger Llama-3-Instruct 70B in obfuscation, content preservation, and grammar. In terms of less content added, \ourmethod still outperforms strong baselines like Llama-3-Instruct 8B, but loses slightly to methods like Gemma-Instruct 7B and Paraphrase; though these methods may be good abstaining from adding new content, this is likely a byproduct of their generations being too succinct and failing to preserve information or obfuscate, as shown by their low human evaluations on these two metrics. 

For overall score, which captures aspects of fluency, content preservation, and obfuscation, \ourmethod performs the best, achieving an overall score of $69.9\%$; the next highest scoring method is Llama-3-Instruct 8B with a score of $66.3\%$, a significant dropoff. Each individual metric must be high to achieve a high product; this indicates that our method on averages produces the obfuscations with the best overall quality, balancing between grammar, content preservation, and obfuscation, rather than optimizing for just one dimension.

\paragraph{Qualitatively, \ourmethod generates more flexible, directed obfuscations compared to other methods} Qualitative results demonstrate that, as designed, \ourmethod provides a strong, personalized obfuscation compared to the more general results of other methods and models. \cref{tab:example_outputs} presents two randomly\footnote{Texts were filtered by a length threshold.} selected texts along with the generations from various models and methods. Consistent with previous work \cite{fisher2024jamdec}, the rule-based methods (MT and Stylo result in poor grammar or loss of content. Conversely, methods based on LLMs tend to maintain grammar and content preservation more effectively.

The most significant difference is evident in the \textit{style} of the generated text. Other methods sometimes struggle to clearly obfuscate and instead generally mimic the original author's style or default to a more formal ``model''-like writing style. In contrast, \ourmethod stands out by providing a more personalized and targeted obfuscation. For instance, in the Blog example (top), \ourmethod generates text that is more formal, uses higher-grade level language, and is longer compared to the original text. Meanwhile, in the Speech example (bottom), it adopts a more sarcastic, less formal tone, and incorporates more function words. 

We also find that this multi-style mixture approach often results in noticeably different sentence structures and punctuation. For example, in the speech text (bottom), the order of the first sentence is reversed compared to the original, a feature not observed in any other generation. Additional generations are available in \cref{appx:more_qual_examples}.

To further highlight the steerability of \ourmethod, we display a randomly selected text from \ourdataset-Speech and random generations created using an adapter in the optimal steering direction\footnote{The optimal direction is calculated based on the automatic style selection method in \cref{sect:stage_2_obf}} for each of the seven style axes in \cref{fig:qual_steerability}. Each generation demonstrates how the choice of adapter significantly transforms the text and influences the type of obfuscation.

\begin{table}[t]
\begin{adjustbox}{width=\linewidth}
\begin{tabular}{lccccccc}
\toprule
\textbf{\# Styles} & \textbf{1} & \textbf{2} & \textbf{3} & \textbf{4} & \textbf{5} & \textbf{6} & \textbf{7} \\
\midrule
Speeches              & 17.0       & 17.7       & \textbf{21.2}       & 19.2       & 6.0        & 17.0      & 11.4       \\
Novels                & 8.6        & 11.2       & 13.0       & 14.4       & 16.3      & 11.2       & \textbf{21.8}       \\
Scholar            & 1.1        & 1.8        & 2.3        & 3.4        & 0.8        & 6.0       & \textbf{16.9 }      \\
Blog                  & 13.1       & 16.5       &\textbf{ 19.6  }     & 18.9       & 12.1       & 10.5       & 6.4   \\ \bottomrule
\end{tabular}
    \end{adjustbox}
\caption{Overall task score on the base adapter merging method using different number of style adapters. We compare the overall task score using $1-7$ style adapters. For all automatic evaluations see \cref{tab:n_style_exploration_all}}
\label{tab:auto_eval_n_style}
\end{table} 
\subsection{Ablations and Other Studies}
\paragraph{Our automatic method of style selection results in better obfuscation than random selection} Although \ourmethod can be used with any arbitrary method of choosing the style axes to change, we do find that choosing based on difference between the average style vector and the author vector improves obfuscation on average by ~$6\%$ over random selection of the same number of weights. We note that the grammar and content remained about equal. More details can be found in \cref{appx:random_style_selection}

\paragraph{Shuffling style adapters when using \ourmethod-Sequential leads to some variation} For \ourmethod-Sequential we experiment with shuffling the order of the chosen style adapters over $n=3$ random shuffling. We found that the order of the styles does have some effect on the obfuscation drop rate (standard deviation of ~$3\% - 6\%$) but little effect on the grammar or content preservation (standard deviation of ~$1\%-2\%$). This was seen strongly when choosing $3+$ styles and in domains with strong style differences among the authors (Speech and Blog). More details are in \cref{appx:shuffle_style}

\paragraph{Changing $5+$ style axes decreases grammaticality} \cref{tab:auto_eval_n_style} shows how the overall task score changes the number of styles chosen to use the adapter merging method increases. At first, both obfuscation drop rate and overall score steadily increase as we increase the number of style adapters, which corresponds with changing more elements of the original text. However, for \ourdataset Speech, Scholar, and Blog, we see a sudden decrease in overall task score when using 5 styles. Investigating this, we found that using $5+$ style adapters leads to an average of $\sim\!16\%$ decrease in grammar and a $\sim\!5\%$ decrease in overall score. More details can be found in \cref{appx:n_style_changes}. 

\section{Related Work}

\paragraph{Authorship Obfuscation Methods} 
Traditional authorship obfuscation methods leverage stylometric insights, such as author invariant features, to obfuscate texts \cite{stylo_method, mansoorizadeh,xing}. However, these methods have been shown to have issues with grammar and fluency due to their strict rule-based approach \cite{fisher2024jamdec}. 

To reduce this behavior, model-based approaches have been developed, such as Mutant-X, a genetic algorithm which utilizes an internal classifier to iteratively "mutate" a sentence \cite{Mahmood2019AGH}. Later work improves on this with an \emph{ensemble} of classifiers rather than a single one \cite{avengers} or via 
variational autoencoders as the base model to generate deferentially private generations \cite{weggenmann}.  Most recently, \citet{fisher2024jamdec} demonstrate the efficacy of smaller LLMs for authorship obfuscation through over-generation and filtering.
However, this method's reliance on a heavy decoding algorithm to generate diverse candidates makes it impractical. 
Some obfuscation methods have also incorporated authorship information \cite{jones2022robert, Shetty2017A4NTAA}. Although these both showed promise, 
they require extensive training and are only applicable in specific use cases.

\paragraph{Parameter Efficient Learning} 

Parameter-efficient adapters, small modules tuned on top of a frozen large model for effective transfer learning, have been proposed for vision \cite{Rebuffi2017LearningMV} and NLP \cite{Houlsby2019ParameterEfficientTL}. Others have extended these methods by tuning specific layers and embeddings \cite{Li2021PrefixTuningOC, Lester2021ThePO}, or by making the adapter matrices an addition to the original model weights themselves rather than additional injected layers \cite{Hu2021LoRALA, Lu2023InferenceTimePA}.

Adjacent to parameter-efficient training strategies are \textbf{ model merging} techniques, which seek to integrate model knowledge by combining their weights \cite{Matena2021MergingMW}; this is efficient and prevents additional inference cost. Merging has been explored extensively in previous work, to combine diverse, targeted domain models \cite{Jang2023PersonalizedSP, Ram2023RewardedST}, or over the same model trained with different seeds or hyperparameters to improve robustness 
\cite{Wortsman2022ModelSA, Ram2022DiverseWA}. Model merging has even been explored with parameter-efficient adapters like LoRA \cite{Huang2023LoraHubEC}. Other lines of work expand on merging techniques, creating strategies beyond simply averaging model weights. \cite{Yadav2023TIESMergingRI, Stoica2023ZipItMM, Yu2023LanguageMA}.

\paragraph{Controllable Generation} Previous work introduces methods to control the content of a generation \cite{lu-etal-2021-neurologic} or steer the style of the generation \cite{liu-etal-2021-dexperts, Lu2023InferenceTimePA}. However these types of controllable generation are less practical for authorship obfuscation, which requires a steerability of the content and the style.

\paragraph{Style Transfer}
Style transfer techniques have utilized both simple models \cite{10.1145/3544903.3544906} and more advanced machine learning models \cite{jin-etal-2022-deep, Hallinan2023STEERUS}. Most approaches depend on training generation models with a dataset, which can be natural or synthetic \cite{jin-etal-2022-deep}. These methods also interpret "style" in various ways, ranging from comprehensive notions like specific authors (e.g., Shakespeare and Hemingway) \cite{strap} to particular stylistic elements (e.g., formality and sentiment) \cite{Fu_Tan_Peng_Zhao_Yan_2018}. Notably, recent, effective style transfer techniques resemble the approach we use, which involves fine-tuning a LLM \cite{Hallinan2023STEERUS, strap}. 
\section{Conclusion}
In this work, we introduce \ourmethod, a novel and interpretable method for authorship obfuscation. By targeting specific fine-grained stylistic elements and leveraging Low Rank Adaptation (LoRA) modules, \ourmethod provides a more interpretable and controllable approach than existing methods based on large language models or other state-of-the-art techniques while still maintaining excellent performance. We show our new method outperforms a suite of strong, state-of-the-art baselines in four diverse domains overall in both automatic and human evaluation.

Additionally, as part of this work, we release a new authorship attribution dataset, \ourtestdataset. This dataset includes two new domains: presidential speeches and fiction books, which were carefully selected to ensure a high degree of topic matching, thereby enriching the dataset's applicability and depth. Furthermore, we develop \ourstyledataset, a collection of 16 parallel, human-validated datasets spanning various stylistic dimensions, which can be employed in future research to further explore and refine the nuances of stylistic text manipulation. These resources aim to advance the field of authorship obfuscation and provide a solid foundation for subsequent studies. %

\section{Limitations and Ethical Considerations}
One limitation of \ourmethod is the requirement of needing trained LoRA adapters and the corresponding style datasets for their training. This necessitates an additional pre-obfuscation step involving separate style corpi and computational training time. However, this is a one-time expense, and the same style adapters can be utilized for multiple authors. In return, users benefit from a more interpretable method for authorship obfuscation.

During obfuscation, \ourmethod does require a slightly higher computational time and memory due to the extra style LoRA adapters compared to using a finetuned model with instructions. For the sequential version of \ourmethod, the computational time is multiplied by the number of styles. The base adapter merging variation outperforms the sequential version and is also more efficient: the time is only increased by a small amount from merging adapters then adding them to the base model, rather than requiring multiple model forward-passes. In inference, no extra time is added, since LoRA weights are seamlessly added to the original model \cite{Hu2021LoRALA}; see \cref{concat} for further analysis. Finally, we note that for the adapter merging with LoraHub, there is also additional time over the base adapter merging to identify the optimal weights. 

An additional limitation is that \ourmethod is developed and tested only on English. However, we believe that the framework (identifying style axes, training LoRA adapters, evaluating original text on these axes, and perturbing based on these evaluations) could be generalized to any language. As such, we believe our pipeline can be effectively adapted to obfuscate other languages; we leave the exploration of these adaptations to future work.

Lastly, our work also has some potential risks. Though the intention of authorship obfuscation is to protect identities in sensitive situations, there is a possibility that malicious users could misuse our method. We acknowledge this as a potential risk for any authorship obfuscation method, which is inherent when creating these methods. 

\section{Acknowledgements}
This research was supported by the DARPA MCS program 
through NIWC Pacific (N66001-19-2-4031), 
IARPA 2022-22072200003, 
and NSF IFML CCF-2019844. 
\bibliography{main}
\pagebreak
\appendix
\section{Code and Artifacts}
The code for \ourmethod is available on Github at \url{https://github.com/jfisher52/StyleRemix}. In addition, we release a demo of our method at \url{https://huggingface.co/spaces/hallisky/StyleRemix}, which allows for arbitrary inputs and rewrites with multiple LoRA adapters across multiple style axes. 

Finally, we release the rest of our datasets and trained classifiers and LoRA adapters \href{https://huggingface.co/collections/hallisky/authorship-obfuscation-66564c1c1d59bb62eaaf954f}{in a HuggingFace collection}. Notably, the collection includes \ourdataset at \url{https://huggingface.co/datasets/hallisky/AuthorMix} and \ourstyledataset at \url{https://huggingface.co/datasets/hallisky/DiSC}.

\section{Extended Ablations and Other Studies}

\subsection{Random Selection of Styles} \label{appx:random_style_selection}
In \cref{sect:stage_2_obf}, we describe a simple automatic method to select the style axes to change for each author. It requires creating an author vector, which is composed of the ten style axes automatic evaluations, and finding the difference for each author compared to the average vector of all authors in a domain. In order to test the efficacy of our style axes selection method, we compare the results of \ourmethod when selecting the styles axes in this way and randomly (over $n=3$ different seeds). 

\begin{figure*}
    \centering
    \includegraphics[width = 1\linewidth]{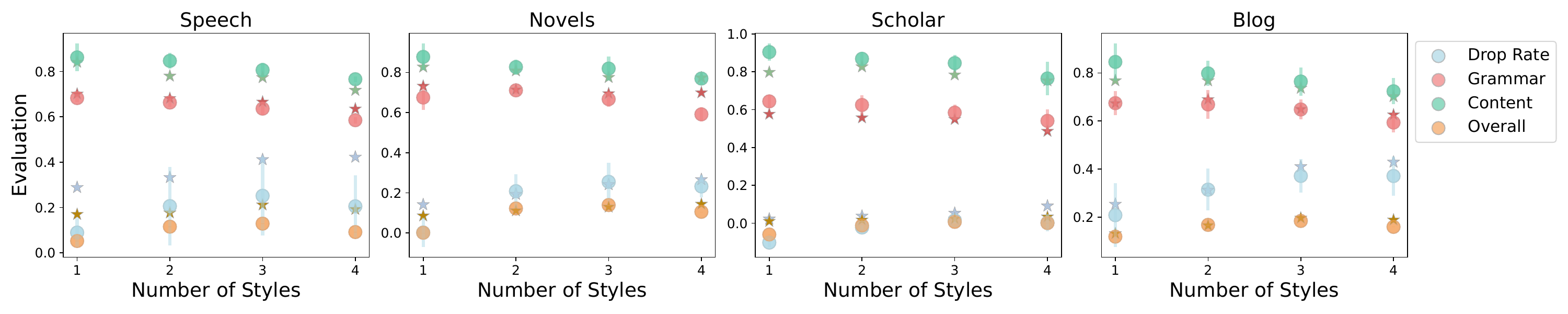}
    \caption{Base model merging with random styles, n = 3}
    \label{fig:base_mm_random}
\end{figure*}

\cref{fig:base_mm_random} shows the average and standard deviation of the drop rate, grammar score, content preservation score and overall task score for each domain randomly choosing $1-4$ styles (circles) and using our automatic method of style axes selection (stars). First, we notice that overall, the grammar and content preservation is mostly similar for both random and the automatic method. However, we do see a large difference in obfuscation drop rate, especially in speech (~$18\%$ average) and Scholar (~$8$ average). These datasets have more modern, similar styles, which might need a more targeted obfuscation rather than the novels (which are written in older English) and the blog (which are very informal). 

\subsection{Comparing with Different LLMs}\label{appx:compare_more_llms}
\begin{table}[!ht]
\begin{adjustbox}{width=\linewidth}
\begin{tabular}{lccccc}
    \toprule
          ~ & \textbf{Mistral}  & \textbf{Gemma}   & \multicolumn{3}{c}{\textbf{\ourmethod}} \\ 
        & V2 &  2B & Seq. & AM & AM + LoraHub$^+$ \\\midrule
                 \rowcolor{gray!25}
        \multicolumn{6}{c}{\ourdataset-\textbf{Speech}} \\ \midrule
\textbf{Drop Rate} & 25.8        & 24.7      & 34.9           & 41.2     &  31.4\\
\textbf{Grammar}   & 67.6        & 70.6      & 61.7           & 66.5     &  63.9\\
\textbf{Content}   & 81.0        & 78.2      & 71.3           & 77.3     &  73.9\\
\textbf{Overall}   & 14.1        & 13.6      & 15.3           & 21.2     &  14.8\\ \midrule
\rowcolor{gray!25}
        \multicolumn{6}{c}{\ourdataset-\textbf{Novels}} \\
\textbf{Drop Rate} & 12.0        & 13.5      & 19.3           & 28.6     &  35.6\\
\textbf{Grammar}   & 69.7        & 72.2      &72.6           & 68.1     &  63.5 \\
\textbf{Content}   & 80.1        & 78.2      & 83.7           & 76.1     &  72.9\\
\textbf{Overall}   & 6.7         & 7.6       &11.8           & 14.8     &  16.5\\ \midrule

 \rowcolor{gray!25}
         \multicolumn{6}{c}{\ourdataset-\textbf{Scholar}} \\
\textbf{Drop Rate} & 0.8         & 1.5       & 1.8            & 9.2      &  11.5\\
\textbf{Grammar}   & 66.8        & 69.5      & 65.8           & 48.6     &  44.7\\
\textbf{Content}   & 88.9        & 87.3      & 78.0           & 75.3     &  68.8\\
\textbf{Overall}   & 2.3         & 2.8       & 3.6            & 3.4      &  3.5\\ \midrule
                 \rowcolor{gray!25}
        \multicolumn{6}{c}{\ourdataset-\textbf{Blog}} \\
\textbf{Drop Rate} & 23.7        & 21.9      & 34.4           & 41.0     &  42.0\\
\textbf{Grammar}   & 68.3        & 71.3      & 66.7           & 64.9     &  65.3\\
\textbf{Content}   & 78.3        & 77.1      & 72.1           & 73.7     &  74.2\\
\textbf{Overall}   & 12.7        & 12.0      & 16.5           & 19.6     &  20.4\\ \bottomrule
    \end{tabular}
    \end{adjustbox}
    \caption{Results of automatic evaluation on other LLMs and methods compared to \ourmethod.}
    \label{tab:auto_eval_other_models}
\end{table}

 For the main experiment we showed the comparison with different like-sized LLMs. Here we provide more comparisons with Mistral V2 \cite{jiang2023mistral} and Gemma (2B) \cite{gemmateam2024gemma} to the three variations of \ourmethod. We show results for all three criteria as well as the overall task score. We see continue to have the highest overall and obfuscation rate compared to these models.

\subsection{Shuffling Styles using the Sequential Method} \label{appx:shuffle_style}
One version of \ourmethod described in \cref{sect:stage_2_obf} is the sequential method, which runs the original text through each adapter sequentially. We hypothesized that the order in which the styles were rewritten might affect the final outcome. To test this, we randomly shuffled the order of the adapters of the styles axes over $n=3$ different seeds when changing $2-4$ styles and tested automatic evaluations as we did in the main experiment. 

\begin{figure*}
    \centering
    \includegraphics[width = 1\linewidth]{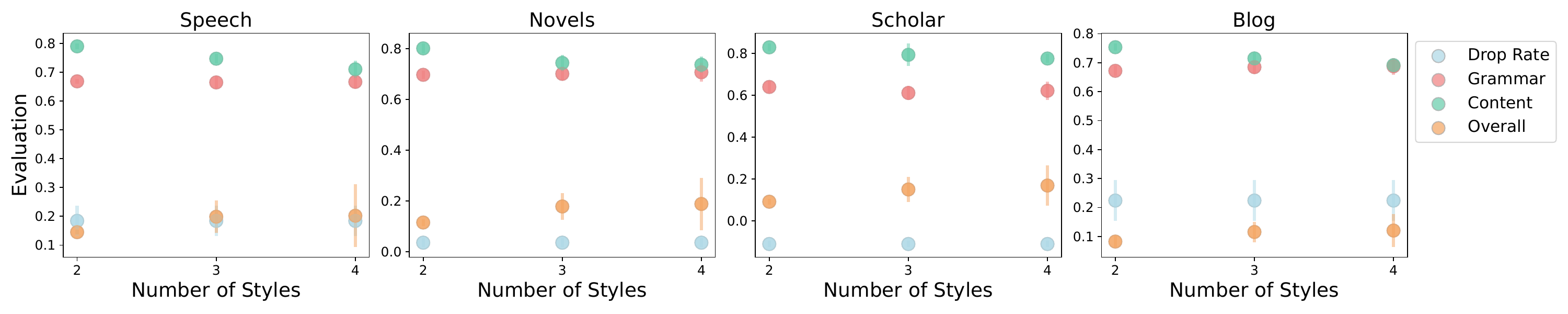}
    \caption{Seq. shuffle n = 3}
    \label{fig:seq_shuffle}
\end{figure*}

\cref{fig:seq_shuffle} shows the average and standard deviation for all the automatic evaluations for each domain and different number of styles changed. We first note that grammar and content preservation remains similar, with very low standard deviation. However, for specific domains, the obfuscation drop rate has a large variation between the three random shuffles. This most diverse obfuscation drop rates seen in Speech ($\sim\!14\%$ standard deviation) and Blog ($\sim\!9\%$ standard deviation). This indicates that the order of adapter in the sequential method could contribute to the overall efficacy of the method. Future work could experiment more with these findings.

\cref{fig:seq_shuffle} shows the automatic evaluations when we shuffle $2-4$ style axes adapters.

\subsection{Number of Styles Change} \label{appx:n_style_changes} In \ourmethod the user can decide how many style adapters to use during obfuscation. We tested how obfuscation drop rate, grammar, and content preservation is affected when more style adapter are added. For this experiment, we used the base model adapter method and selected $1-7$ styles using the difference from the author vector to the average domain vector. 

\cref{tab:n_style_exploration_all} shows all the automatic evaluations for each number of style. At first, we see a stead increase in both obfuscation drop rate and overall score as we increase style adapters. This corresponds with changing more elements of the original text. However, as mentioned in our main paper, we see on average a ~$5\%$ decrease in overall task score when using 4 to 5 style adapters. Then, as the number of style adapter increase, we see a stead decrease in content preservation and grammar. This correlates with a qualitative decrease in generations seen as we increase the styles over 5.  
\begin{table}[]
\begin{adjustbox}{width=1\linewidth}

\begin{tabular}{lrrrrrrr}
\textbf{\# of Styles} & \multicolumn{1}{l}{\textbf{1}} & \multicolumn{1}{l}{\textbf{2}} & \multicolumn{1}{l}{\textbf{3}} & \multicolumn{1}{l}{\textbf{4}} & \multicolumn{1}{l}{\textbf{5}} & \multicolumn{1}{l}{\textbf{6}} & \multicolumn{1}{l}{\textbf{7}} \\ \midrule
              \rowcolor{gray!25}
        \multicolumn{8}{c}{\ourdataset-\textbf{Speech}} \\
\textbf{Drop Rate}    & 28.9  & 33.3  & 41.2  & 42.3  & 13.6  & 47.4  & 44.6  \\
\textbf{Grammar}      & 70.0  & 68.1  & 66.5  & 63.5  & 61.4  & 52.9  & 46.1  \\
\textbf{Content}      & 84.1  & 78.0  & 77.3  & 71.7  & 72.2  & 67.7  & 55.4  \\
\textbf{Overall}      & 17.0  & 17.7  & 21.2  & 19.2  & 6.0   & 17.0  & 11.4  \\
              \midrule
              \rowcolor{gray!25}
        \multicolumn{8}{c}{\ourdataset-\textbf{Novels}} \\
\textbf{Drop Rate}    & 14.2  & 19.3  & 24.2  & 26.7  & 36.1  & 32.9  & 83.7  \\
\textbf{Grammar}      & 73.2  & 71.4  & 69.5  & 69.9  & 61.3  & 49.8  & 50.4  \\
\textbf{Content}      & 82.7  & 80.9  & 77.6  & 77.4  & 73.7  & 68.4  & 51.6  \\
\textbf{Overall}      & 8.6   & 11.2  & 13.0  & 14.4  & 16.3  & 11.2  & 21.8  \\
              \midrule
              \rowcolor{gray!25}
        \multicolumn{8}{c}{\ourdataset-\textbf{Scholar}} \\
\textbf{Drop Rate}    & 2.3   & 3.8   & 5.3   & 9.2   & 2.3   & 50.4  & 73.5  \\
\textbf{Grammar}      & 57.7  & 55.7  & 54.9  & 48.6  & 48.1  & 38.6  & 48.4  \\
\textbf{Content}      & 79.8  & 82.7  & 78.4  & 75.3  & 71.7  & 30.8  & 47.5  \\

\textbf{Overall}      & 1.1   & 1.8   & 2.3   & 3.4   & 0.8   & 6.0   & 16.9  \\
              \midrule
              \rowcolor{gray!25}
        \multicolumn{8}{c}{\ourdataset-\textbf{Blog}} \\
\textbf{Drop Rate}    & 25.4  & 31.2  & 41.0  & 42.9  & 34.3  & 38.3  & 41.4  \\
\textbf{Grammar}      & 67.3  & 68.9  & 64.9  & 62.6  & 55.3  & 46.4  & 44.0  \\
\textbf{Content}      & 76.8  & 76.8  & 73.7  & 70.2  & 63.7  & 59.1  & 35.3  \\
\textbf{Overall}      & 13.1  & 16.5  & 19.6  & 18.9  & 12.1  & 10.5  & 6.4   \\ \bottomrule         
\end{tabular}
\end{adjustbox}
\caption{Results of automatic evaluation on the base adapter merging method using different number of style adapters. We show the obfuscation drop rate, grammar, content preservation, and overall task score using $1-7$ style adapters. }
\label{tab:n_style_exploration_all}
\end{table} 
\subsection{Author Style Vector Analysis}\label{appx:author_style_vector_analysis}
In the pre-obfuscation phase, we choose 7 specific style axes to train the LoRA adapters; length, use of function words, grade level, voice, use of sarcasm, formality, and writing intent. Some of these style axes have rule-based evaluations, and others have classifier-based evaluations. We used these automatic evaluations to create a unique author vector for each author in a domain and use the difference in this vector compared to other authors in the same domain to choose the styles axes to change during obfuscation. Although these selected style axes are just a subsample of suitable options, we wanted to explore how well these author vectors separate the authors in our test data set.

To analyze this, we first created an author vector for each author by taking the average of each automatic evaluation over the paragraphs in the authors test set. This resulted in 14 (authors), vectors with 7 (style axes) entries each. We then performed a principle component analysis (PCA) to reduce the size of the vector dimension to explain at least $90\%$ of the variance in the data (it went from 7 to 4 dimensions). We note that the first two dimensions account for ~$70\%$ of the variation. 

\begin{figure}
    \centering
    \includegraphics[width = 1\linewidth]{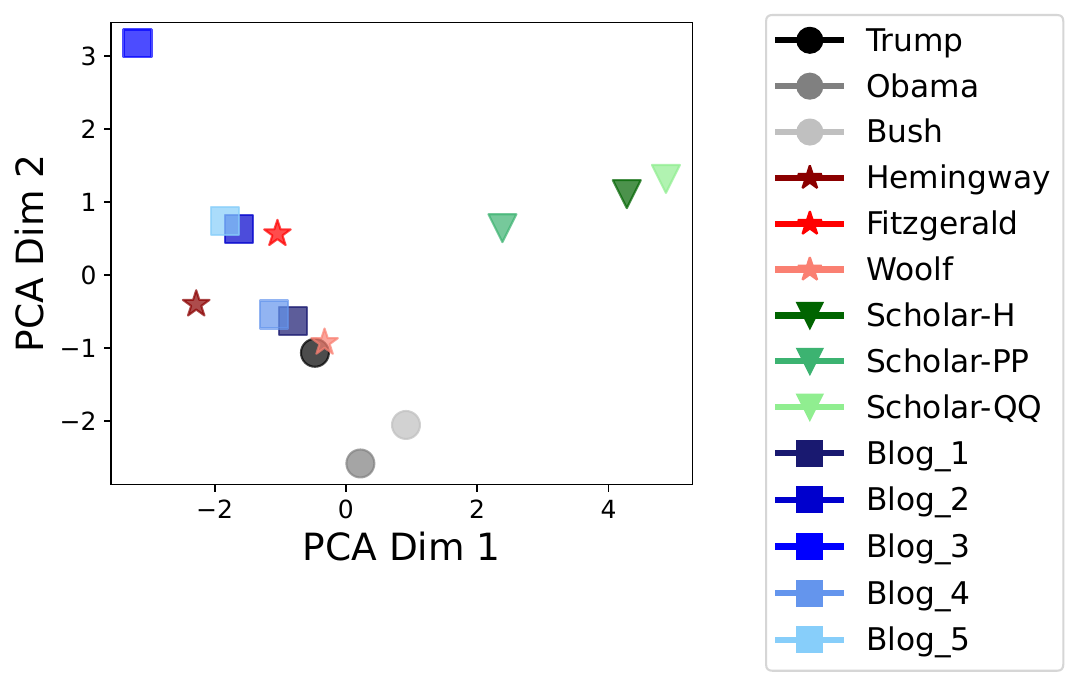}
    \caption{PCA, clustering analysis}
    \label{fig:style_clustering_analysis}
\end{figure}

\cref{fig:style_clustering_analysis} compares all the authors (across all domains) using the first and second component of the PCA. First, we notice that the Scholar (triangles) and Speech (circle) domains have distinct clusters away from the other two domains. The most spread out domain is Blog (square) with one author quite different from the rest. Lastly, we see that the novel (start) dataset is closely clustered together, but are quite similar to 4 of the blog authors. We note that four of the blog authors have more story-telling writing styles, while the last one has a more diary-like, very informal writing style. This seems consistent then that it would cluster similarly as novels. 

Overall, this analysis showed starting evidence that our style axes vectors were able to separate the diverse writing styles. Future research could continue to explore the types of style axes that are most important when obfuscating.  

\subsection{More Qualitative Examples}\label{appx:more_qual_examples}
In \cref{tab:qual_examples} we provide more examples from each author in the \ourdataset. We note that we selected these samples by \textit{randomly} selecting $3$, paragraphs of less than $45$ words for each author and then selecting the example from these three. For \ourmethod, we used the base model adapter method with $3$ style adapters. From these examples, especially the Blog and Novels, we see the qualitative benefits of \ourmethod and it's flexibility to adapt to different original author styles. 

\subsection{Tradeoff between Obfuscation, Content Preservation, and Grammar}
We want to note that there is a natural trade-off between authorship obfuscation, content preservation, and grammar. For example, a naive copying baseline would have high grammar and perfect content preservation but low obfuscation. On the other extreme, a complete gibberish output would score very low on grammar and content but high on obfuscation. 

This phenomenon is well-documented in the context of style transfer. To assess the overall quality of generations where there are multiple objectives, previous work in style transfer \cite{strap, Hallinan2023STEERUS, xu-etal-2018-unpaired, patel2023lowresourceauthorshipstyletransfer} proposes taking the \textit{product} (or geometric mean) of the metrics (instead of drop rate, for style transfer we have target style strength, still bounded). The intuition is that these style transfer systems should \textit{ jointly optimize all metrics} rather than just one or two; this is reflected in taking a product.

In line with this past style transfer work, we choose to use the product, as we think high-quality obfuscations should jointly prioritize the three metrics of fluency, meaning similarity, and obfuscation, and so that we do not encourage systems that only optimize one or two of these metrics.
However, we note that certain aspects might be more important for users than others, and they might not want to use an equally weighted total. 

 However, as an alternative overall metric, \textit{we present the overall score as an equally weighted average} of the drop rate, grammar, and content score below, rather than a product. See these results in \cref{tab:average_for_overall}. We note that this was used in other authorship obfuscation papers as an overall total metric as well \cite{fisher2024jamdec}. Here, we again see \textbf{\ourmethod performs best overall in 3 of the 4 datasets}. Note that the decrease in performance on Scholar split of \ourdataset is due to the very low obfuscation rate among all methods, which results in only a difference of $5\%$ between \ourmethod and the main method.

\begin{table*}[]
\begin{adjustbox}{width=1\linewidth}
\begin{tabular}{llllllllllllll}
\toprule
                    &   & \multicolumn{2}{l}{\textbf{Llama-2-Chat}} & \multicolumn{2}{l}{\textbf{Llama-3-Inst}} & \textbf{Gemma-Inst} & \textbf{Paraphrase} & \textbf{MT} & \textbf{Stylo} & \textbf{JD} & \multicolumn{3}{c}{\textbf{StyleRemix}}          \\
                 
\textbf{Dataset} & & 7B                  & 13B                 & 8B                  & 70B                 & 7B                  &                     &             &                &             & Seq.           & AM             & AM+LoraHub$^+$    \\
\midrule
Speech           & & 56.60               & 57.30               & 55.17               & 55.73               & 56.50               & 59.73               & 51.43       & 47.47          & 47.43       & 55.97          & \textbf{61.67} & 56.40          \\
Novels           &  & 55.63               & 56.07               & 55.13               & 56.07               & 55.07               & 53.47               & 46.17       & 46.13          & 48.23       & \textbf{58.53} & 57.60          & 57.33          \\
Scholar          & & 52.27               & 52.03               & 51.53               & 51.03               & 51.40               & \textbf{53.73}      & 49.60       & 40.47          & 43.00       & 48.53          & 44.37          & 41.67          \\
Blog             & & 56.20               & 56.47               & 57.07               & 56.90               & 58.10               & 57.47               & 45.00       & 42.33          & 54.03       & 57.73          & 59.87          & \textbf{60.50} \\ \bottomrule
\end{tabular}
\end{adjustbox}
\caption{Results from \cref{tab:auto_eval_baseline}, but now using an \textbf{equal-weighted average} as the overall score rather than the product of metrics, across all domains and models.}
\label{tab:average_for_overall}
\end{table*} 

\section{Method Details}\label{appx:method_details}
\subsection{Style Axes Selection and Evaluation} \label{appx:style_data_eval}
We choose seven different style axes. The first three style axes have rule-based evaluation; length, use of function words, and grade level. For length, we evaluate using the average words per sentence and for function words we use the number of function words. Additionally, we incorporated "grade level," which primarily measures the number of syllables. Since this measure can vary slightly, we averaged three similar metrics: the Flesch-Kincaid (FK; \citealp{flesch1948new}), Linsear Write (L; \citealp{linsear}), and the Gunning Fog Index (GF; \citealp{gunning1952technique}) metrics. The exact formulas are given below; for more details, see \href{https://github.com/textstat/textstat}{https://github.com/textstat/textstat}.
\begin{itemize}
\item \textbf{Flesch-Kincaid} is computed via:
\begin{align*}
\resizebox{.8\hsize}{!}{$KF = 0.39\left(\frac{\text{total words}}{\text{total sentences}}\right) + 11.8\left(\frac{\text{total syllables}}{\text{total words}}\right) - 15.59$}
\end{align*}
\item \textbf{Linsear Write} is computed by:
\begin{enumerate}
    \item Take a 100-word sample from the text
    \item Make a score starting with 0. For every ``easy'' word ($\leq$ 2 syllables), add 1 point. Otherwise add 3 points (``hard'' words have $\geq$ 3 syllables).
    \item Divide points by number of sentences in the 100-word sample.
    \item Divide by 2 if the points $< 20$, otherwise divide by 2 and subtract 1.
\end{enumerate}
\item \textbf{Gunning Fog} is computed by selecting a passage around 100-words long, then applying the following formula:
\begin{align*}
\resizebox{.8\hsize}{!}{$GF = 0.4\left[\left({\frac {\mbox{words}}{\mbox{sentences}}}\right)+100\left({\frac {\mbox{complex words}}{\mbox{words}}}\right)\right]$}
\end{align*}
where complex words are words with three or more syllables.
\end{itemize}

The next four style axes have model-based evaluation; use of sarcasm, voice (active or passive), formality, and writing intent (descriptive, expository, narrative, and persuasive). Although these were chosen arbitrarily, we believe they do reflect some unique aspects of authorship style. However, these styles do require a unique classifier to automatically evaluate a text. For formality we used a RoBERTa-base \cite{Liu2019RoBERTaAR} based formality classifier \cite{formality_class}, found at \href{https://huggingface.co/s-nlp/roberta-base-formality-ranker}{https://huggingface.co/s-nlp/roberta-base-formality-ranker}.

However, for the other three axes (voice, sarcasm, and writing intent) there was not a reliable, existing model, so we trained our own classifiers. We follow the same procedure to make \ourstyledataset detailed in \cref{preobf}, but 1) with \emph{different} base training data, to ensure that there is no overlap between the classifier and adapter data and 2) only for the following style elements: voice passive, voice active, sarcasm less, sarcasm more, and persuasive, expository, narrative, and descriptive. With the new datasets of length 1500 for each style element, we then train RoBERTA-large \cite{Liu2019RoBERTaAR} discriminators for the voice, sarcasm, and writing intent categories, splitting the train into 85\% train and 15\% dev set. We set the seed to 0 and train with a batch size of 128, learning rate of 5e-5, and for 5 epochs. 

For all models, we choose the checkpoint with the best evaluation accuracy product (to ensure high accuracy for all classes); this corresponded to 100\%, 99.1\%, 45.5\% for sarcasm, voice, and type respectively. Each model took approximately 1 hour to train using 1 NVIDIA A100 GPU with 80 GB of VRAM.

\subsection{\ourstyledataset Training Data and Evaluations} \label{appx:training_data_details}
We use GPT4-Turbo \cite{openai2023gpt4turbo} to distill the style axes into 16 parallel training sets. We collect 1500 paragraphs from Wikipedia, books and plays, and blogs, then prompt GPT4 with the following: ``Rewrite the following paragraph to include the same content but {specific prompt}\textbackslash n Paragraph: {paragraph} \textbackslash n Rewrite: `` 
where paragraph is the original data. Depending on the target style, we change the specific prompt to:
\begin{itemize}
\item \textbf{Length short}: ``being more succint''
\item \textbf{Length long}: `` being more verbose.''
\item \textbf{Lower grade-level}: ``using language an early elementary school student can understand.''
\item \textbf{Higher grade-level}: ``use high school reading level or above.''
\item \textbf{More function words}: ``using far less function words (i.e. pronouns, determiners, and conjunctions).''
\item \textbf{Less function words}: ``using far more function words (i.e. pronouns, determiners, and conjunctions).''
\item \textbf{More sarcasm}: ``with more sarcasm.''
\item \textbf{Less sarcasm}: ``with less sarcasm.''
\item \textbf{More formal}: ``with more formal language.''
\item \textbf{More informal}: ``with more formal language.''
\item \textbf{Active voice}: ``with active voice.''
\item \textbf{Passive voice}: ``with passive voice.''
\item \textbf{Persuasive writing style}: ``with persuasive writing style.''
\item \textbf{Expository writing style}: ``with expository writing style.''
\item \textbf{Narrative writing style}: ``with narrative writing style.''
\item \textbf{Descriptive writing style}: ``with descriptive writing style.''
\end{itemize}
We use sampling with a temperature of 1.0. As a result of this prompting, we achieve $1500 \cdot 16 = 24000$ generations spanning $16$ unique style directions from GPT-4.

We then validate the quality of this data. For axes with available automatic metrics, specifically length, function words, grade level, and formality, we run their respective metrics on the original texts, and on the GPT-4 generations in both directions, ie., we run the formality classifier on the original texts, and on both the more and less formal GPT-4 generations. For the axes without automatic evaluation, we instead randomly evaluate 10\% of them. Specifically, we randomly combine generated data from the same style axis but different directions (such as more and less sarcasm), and ask annotators (three NLP experts) to label if the style axis is high or low (or the specific type for Writing Type), then compute the accuracy.

\cref{tab:train_data_eval} shows the results. For the metrics that we can automatically evaluate, our generated data captures the desired axes and directions well; for example, the texts steered towards higher length have the highest average number of words per sentence. For sarcasm and voice, human evaluations of 97.7\% and 93.7\% respectively indicate that the generations match the targeted directions. For writing intent, the human evaluation accuracy is 77.7\% which is still a good number as the task of discriminating between four classes is inherently more complex.

\subsection{Style Adapter Training}
\label{lora}
We train LoRA adapters \cite{Hu2021LoRALA} using each of the 16 generated parallel datasets. Specifically, we train LLama-3 8B (base model) on the following prompt for each of the datasets: 

\texttt{\textless bos\textgreater \#\#\# Original:\{original\} \textbackslash n \#\#\# Rewrite: \{rewrite\} \textless eos\textgreater}

where original and rewrite denote the original text and rewrite is the text we generated from GPT-4. Note that the format we train on is the same for all parallel datasets to make future model merging more effective.

We train the 16 LoRA modules each for 5 epochs with a seed of 0, batch size of 6, and a max sequence length of 512; we choose the checkpoint with the best eval loss and have an early stopping criteria of 5. For LoRA parameters, we use use $r=32$, the rank of the matrix, and the alpha and dropout values of 32 and 0.01 respectively. Overall, each LoRA adapter involves training 13 million parameters each, about 0.16\% of the total parameters in LLama-3 8B.

All of our models train well over time on both train and eval loss; please see our repository for exact training curves and loss numbers for the 16 models. We train each of the models on a single A100 80GB GPU for about 2 hours each.

\subsection{Concatenating Style Adapters}
\label{concat}

Given a model with weights $W_0$ of dimension $d \times d$ LoRA freezes $W_0$ and trains two matrices: $A$ of size $r \times d$ and $B$ of size $d \times r$. At inference, we use the new weights of $W = W_0 + BA$.

In the situation when we have $n$ LoRA adapaters, parameterized by $A_1...A_n$ and $B_1...B_n$ and want to ensemble them for inference, we use \textbf{concatenation}. Specifically,  we concatenate each of the $A_1...A_n$ matrices resulting in a matrix $A_{1...n}$ of size $nr \times d$. Similarily, we concatenate each of $B_1...B_n$ resulting in a matrix $B_{1...n}$ of size $d \times nr$. We then can combine the matrices the same way to get new weights of $W = W_0 + B_{1...n}A_{1...n}$. Notably, we have no additional inference latency by concatenating the vectors, only a slightly increased fixed matrix multiplication cost.

\subsection{Style and Weight Selection}\label{appx:our_method_style_weight_selection}
As described in our paper, we developed an automatic method for selecting the style axes to change, direction, and weights of the adapters. First, we create an author vector for each author in a domain, which is a vector with $10$ automatic evaluations; average words/sentences, average number of function words, average grade level (using FK, L, GF) \cite{flesch1948new, linsear, gunning1952technique}, average likelihood score from formality classifier \cite{formality_class}, average likelihood score from sarcasm classifier (see \cref{appx:style_data_eval} for more details), average likelihood from a voice classifier (see \cref{appx:style_data_eval} for more details), average classification into each of the four writing intents. We label this vector for author $i$ as \textbf{$x_i$}$ \in \mathbb{R}^{10}$.

In order to select the $k$ number of styles axes to change, we use the other authors in the same domain as a baseline. Specifically, we average the values from all authors in the domain and find the styles of author $i$ that are furthest from this average vector. More specifically, we use the following formula:
\begin{align*}
    \text{styles to change} = \text{top}_k\left(\left\vert x_i - \sum_{j=1}^m x_j \right\vert\right),
\end{align*}
where we have $m$ total authors in the domain and $top_k($\textbf{$y$}$)$ is a function which selects the rows of $y$ with the highest values. Similarly, we use the sign of this difference to decide on the direction of the change. For example, if the sign of the difference is negative, then the author's style value is lower than the average and we will implement a higher direction (driving the style up to average). 

Once the styles axes are selected, we use different methods for choosing the adapter weights for each style axes.  First, we also use the author difference vector to select the weight of the adapter. To do this, we calculate the number of standard deviation the author's value is from the average vector. We then use this metric to map to a static weight; see \cref{tab:std_to_adapter_weight_mapping}. We note that these weights were selected in line with past work \cite{Huang2023LoraHubEC}.

Second, we employ a non-gradient based optimization method called LoraHub \cite{Huang2023LoraHubEC}. This method uses a few validation examples to optimize the values. For this method, we developed our our loss function which is the sum of the chosen style axes automatic evaluations as well as the grammar. Specifically,
\begin{align*}
 L = \sum_{v_i \in \text{selected axes}}\begin{cases} 
      v_i & v_i \leq \frac{1}{m} \sum_{j=1}^m x_i\\
      1 - v_i & v_i > \frac{1}{m} \sum_{j=1}^m x_i
   \end{cases}
\end{align*}
where $v_i$ represents the style value for a selected style axis of the obfuscated text and the grammar score. In \cref{tab:base_vs_lorahub_weights}, we show the difference between the base initial weights, chosen using the static method, to the once optimized using LoraHub. 
\begin{table}[ht]
\centering
\footnotesize
\begin{tabular}{ c c }
\toprule
\textbf{\# of Std. Deviations} & \textbf{Adapter Weight}\\
\midrule
0-1 & 0.7 \\
1-2 & 0.9 \\
2-3 &  1.2\\
3+ & 1.5 \\
\bottomrule
\end{tabular}
\caption{This shows the static mapping used in the base adapter merging method. We use the number of standard of deviations an authors automatic style score is from the average style score of all authors in that domain. The static values were chosen base on past work \cite{Huang2023LoraHubEC}}
\label{tab:std_to_adapter_weight_mapping}
\end{table} 
\begin{table*}[]
\begin{adjustbox}{width=1\linewidth}
\begin{tabular}{llll}
\textbf{Author}     & \textbf{Styles Axes}                                                                                                            & \textbf{Base Weights} & \textbf{LoraHub Weights}      \\
\toprule

                 \rowcolor{gray!25}
        \multicolumn{4}{c}{\textbf{3 Style Adapters}} \\ \midrule

\textbf{Trump}      & {[}'grade level', 'length', 'sarcasm'{]}                                           & {[}0.9, 0.9, 0.9{]}       & {[}1.18, 0.96, 0.91{]}   \\
\textbf{Obama}      & {[}'length', 'sarcasm', 'persuasive'{]}       & {[}0.7, 0.7, 0.7{]}       & {[}0.68, 0.74, 0.75{]}   \\
\textbf{Bush}       & {[}'sarcasm', 'formal', 'grade level'{]}                                               & {[}0.7, 0.7, 0.7{]}       & {[}0.71, 0.56, 0.55{]}   \\
\textbf{Hemingway}  & {[}'grade level', 'sarcasm', 'expository'{]}   & {[}0.9, 0.9, 0.7{]}       & {[}1.16, 0.91, 0.71{]}   \\
\textbf{Fitzgerald} & {[}'descriptive', 'grade level', 'sarcasm'{]}  & {[}0.7, 0.7, 0.7{]}       & {[}0.65, 0.58, 0.41{]}   \\
\textbf{Woolf}      & {[}'expository', 'formal', 'grade level'{]}    & {[}0.9, 0.7, 0.7{]}       & {[}1.17, 0.64, 0.95{]}   \\
\textbf{Scholar-H}  & {[}'descriptive', 'voice', 'sarcasm'{]}            & {[}1.5, 0.7, 0.9{]}       & {[}0.92, 0.28, 0.64{]}   \\
\textbf{Scholar-PP} & {[}'descriptive', 'grade level', 'voice'{]}    & {[}1.5, 0.7, 0.9{]}       & {[}1.42, 0.72, 0.93{]}   \\
\textbf{Scholar-QQ} & {[}'length', 'grade level', 'narrative'{]} & {[}0.9, 0.9, 1.5{]}       & {[}1.16, 0.90, 1.46{]}   \\
\textbf{Blog-1}     & {[}'expository', 'grade level', 'formal'{]}    & {[}0.9, 0.9, 0.7{]}       & {[}0.90, 0.90, 0.95{]}   \\
\textbf{Blog-2}     & {[}'length', 'expository', 'formal'{]}         & {[}0.7, 0.7, 0.7{]}       & {[}0.93, 0.65, 0.68{]}   \\
\textbf{Blog-3}     & {[}'sarcasm', 'descriptive', 'formal'{]}           & {[}0.9, 0.7, 0.9{]}       & {[}0.78, 0.55, 0.74{]}   \\
\textbf{Blog-4}     & {[}'formal', 'sarcasm', 'narrative'{]}             & {[}0.7, 0.7, 0.7{]}       & {[}0.68, 0.45, 0.67{]}   \\
\textbf{Blog-5}     & {[}'formal', 'voice', 'expository'{]}              & {[}0.7, 0.9, 0.7{]}       & {[}0.61, 0.77, 0.50{]} \\

                \rowcolor{gray!25}
        \multicolumn{4}{c}{\textbf{3 Style Adapters}} \\ \midrule
\textbf{Trump}      & {[}'length', 'grade level', 'persuasive', 'sarcasm'{]} & {[}0.9, 0.9, 0.9, 0.9{]}  & {[}1.27, 1.15, 0.88, 0.85{]}  \\
\textbf{Obama}      & {[}'grade level', 'sarcasm', 'persuasive', 'length'{]} & {[}0.7, 0.7, 0.7, 0.7{]}  & {[}0.70, 0.70, 0.70, 0.70{]}  \\
\textbf{Bush}       & {[}'formal', 'descriptive', 'grade level', 'sarcasm'{]}     & {[}0.7, 0.9, 0.7, 0.7{]}  & {[}0.32, 0.07, 0.34,  1.05{]} \\
\textbf{Hemingway}  & {[}'sarcasm', 'grade level', 'expository', 'length'{]}  & {[}0.9, 0.9, 0.7, 0.9{]}  & {[}0.98, 0.80, 0.66,  0.96{]} \\
\textbf{Fitzgerald} & {[}'sarcasm', 'descriptive', 'grade level', 'length'{]} & {[}0.7, 0.7, 0.7, 0.7{]}  & {[}0.73, 0.70,  0.67,  0.72{]} \\
\textbf{Woolf}      & {[}'length', 'grade level', 'formal', 'narrative'{]}    & {[}0.7, 0.7, 0.7, 0.9{]}  & {[}0.06, 0.30, 0.77,  0.24{]} \\
\textbf{Scholar-H}  & {[}'sarcasm', 'expository', 'voice', 'formal'{]}                & {[}0.9, 1.5, 0.7, 0.7{]}  & {[}1.44, 1.36, 0.60,  0.74{]} \\
\textbf{Scholar-PP} & {[}'formal', 'grade level', 'descriptive', 'voice'{]}       & {[}0.9, 0.7, 1.5, 0.9{]}  & {[}1.24, 0.55, 1.47,  0.59{]} \\
\textbf{Scholar-QQ} & {[}'length', 'narrative', 'formal', 'grade level'{]}    & {[}0.9, 1.5, 0.7, 0.9{]}  & {[}0.91, 1.25, 0.70,  0.90{]} \\
\textbf{Blog-1}     & {[}'formal', 'narrative', 'length', 'grade level'{]}    & {[}0.7, 0.9, 0.9, 0.9{]}  & {[}1.07, 1.16, 0.80,  0.67{]} \\
\textbf{Blog-2}     & {[}'expository', 'length', 'formal', 'sarcasm'{]}           & {[}0.7, 0.7, 0.7, 0.7{]}  & {[}0.76, 0.70, 0.71,  0.66{]} \\
\textbf{Blog-3}     & {[}'formal', 'grade level', 'sarcasm', 'descriptive'{]}     & {[}0.9, 0.9, 0.9, 0.7{]}  & {[}1.15, 0.90, 0.90,  0.70{]} \\
\textbf{Blog-4}     & {[}'narrative', 'formal', 'sarcasm', 'length'{]}            & {[}0.7, 0.7, 0.7, 0.7{]}  & {[}0.58, 0.28, 0.46, 0.95{]} \\
\textbf{Blog-5}     & {[}'descriptive', 'voice', 'grade level', 'formal'{]}       & {[}0.7, 0.9, 0.7, 0.7{]}  & {[}0.69, 0.70, 0.58,  0.59{]}\\ \bottomrule
\end{tabular}
\end{adjustbox}
\caption{Comparison of the initial base weights, chosen using the standard deviation to static mapping, and the optimized LoraHub weights, found using our customized loss function. We show the style axes changed, the base weights and the LoraHub weights for each author in each domain. }
\label{tab:base_vs_lorahub_weights}
\end{table*}

\section{Experimental Details}\label{appx:experimental_details}
In this section we provide full details of the experimentation used in this paper. We start with the dataset in Appendix \ref{appx:exp_data}, method implementations for each method in Appendix \ref{appx:exp_methods_implementation}, and our evaluation methodology in Appendix \ref{appx:exp_eval}.

\subsection{Software} We used Python 3.10.13, Pytorch 2.1.2, HuggingFace Transformers 4.39.3. and NLTK 3.8.1. All code is licensed under the
Apache License 2.0.

\subsection{Hardware} All experiments were run on eitiher a single NVIDIA A100 GPU or 4 NIVIDIA A100 GPUs with 80GB memory. We estimate our total computational use to be approximately 80 GPU hours.

\subsection{Data}\label{appx:exp_data}
\begin{table}[ht]
\begin{adjustbox}{width=.5\textwidth}
\centering
\footnotesize
\begin{tabular}{ c c c c c c}
\toprule
\textbf{Dataset} & \textbf{Author} & \textbf{Train} & \textbf{Eval} & \textbf{Test} & \textbf{Total}\\
\midrule
Speeches & Trump & 6,443 & 1,596 &2,677& 10,716\\
& Obama & 810 & 189 & 331& 1,330\\
& Bush & 617 & 139 & 251& 1,007\\
\midrule
Novels & Hemingway & 1,516 & 504 & 504& 2,524\\
& Fitzgerald& 2,658 & 885 & 885 & 4,428\\
& Woolf & 1,469 & 488 & 488 & 2,445\\
\midrule
Scholarly & H & 91 & - & 45&  136\\
& PP & 110 & - & 85&  195\\
& QQ & 85 & - & 67& 152\\
\midrule
Blog & 1 & 3,399 & - & 677&4,076 \\ 
& 2 & 1,073 & - & 143& 1,216 \\
& 3 & 1,064 & - & 210&  1,274\\
& 4 & 595 & - & 217&  812\\
& 5 & 396 & - & 142&  538\\
\bottomrule
\end{tabular}
\end{adjustbox}
\caption{Details of \ourdataset, including the number of samples for the test/eval/train for each domain.}
\label{tab:test_dataset_details}
\end{table}
 As mentioned, we wanted to use a test dataset which had a wide range of diverse authorship styles as well as domains. For this reason, we decided to create a new data set on authorship obfuscation called \ourdataset. This dataset is composed for four domains; presidential speeches, early 1900s fiction novels, scholarly articles, and dairy-style blog entries. Altogether, \ourdataset contains over 30,000 high-quality paragraphs from 14 authors.

For the presidential domain, we curate and clean a novel collection of high-quality presidential speeches from George W. Bush ($n=38$), Barack Obama ($n=29$), and Donald Trump ($n=26$)\footnote{These presidents were selected due to their diverse styles but similar time periods, which minimizes content discrepancies.}, transcribed by  the Miller Center \cite{miller_center}\footnote{\url{https://data.millercenter.org}} at the University of Virginia. We broke the speeches naturally into paragraphs and then selected all paragraphs between $2-5$ sentences. This resulted in a total of $n= 13K$ paragraphs. 

Similarly, we also decided to develop a new collection of early 1900s fiction writers from the with strong writing styles, therefore we choose text from books by Ernest Hemingway, F. Scott Fitzgerald, and Virginia Woolf which were collected from Project Gutenberg \cite{Project_Gutenberg}. We selected the top $4$ most popular books on Project Gutenberg for each author and then again, used the natural paragraphs from each author. We selected all paragraphs between $2-5$ sentences. This resulted in a total of $n= 9K$ paragraphs. 

Lastly, we altered the existing data from two current datasets, the Extended-Brennan Greenstad \cite{amt_dataset} which is a collection of ``scholarly'' short (500-word) paragraphs gathered from Amazon Mechanical Turk (AMT) and the Blog Authorship corpus \cite{blog_dataset}, a collection of blogs (diary-style entries) that were posted to blog.com. We note, these datasets match those used in \cite{avengers}, \cite{Mahmood2019AGH}, and \cite{fisher2024jamdec}. For the AMT dataset, we used authors "h", "pp", and "qq" and we artificially created paragraphs by chunking the text into a random collection of 2-5 sentences (as the text is not naturally broken into paragraphs). For the Blog dataset, we used authors "5546", "11518", "25872", "30102", "30407", we used the natural paragraphs. Then, to match the speech and novel domains, we edited to include all paragraphs between $2-5$ sentences and $3$ words. This resulted in $n= 500$ and $n= 8K$ paragraphs for the AMT and Blog accordingly. 

\subsubsection{Artifact Terms of Use} Our artifacts allow for fair use under Project Gutenberg \cite{Project_Gutenberg}: \url{https://gutenberg.org/policy/terms_of_use.html}

\subsection{Method Implementation}\label{appx:exp_methods_implementation}
 
\subsubsection{Baselines}

 For each baseline, we use the optimal set of hyperparameters reported in its respective paper.

\paragraph{Stylometric (Stylo)} We used \cite{stylo_method} method for AO using stylometric methods, which was originally proposed in the PAN-2016 Author Masking Shared Task competition \cite{pan2016_results}. This method calculates metrics for 12 features that are indicative of style, then modifies the text, so these metrics align with an "average" value. The "averages" were calculated using a combination of training sets including the PAN-2016 Author Obfuscation task \cite{pan2016_results} and public domain books from Project Gutenberg \cite{Project_Gutenberg}. Examples of the metrics this method uses include the average number of words per sentence, word frequency, and the use of uppercase letters. Changes employed include actions such as sentence splitting and merging, substitution of words with synonyms, and alterations in spelling. For a full list of metrics and proposed changes, see the \cite{stylo_method}. To further enhance the obfuscation process, the method introduces "noise" by modifying words that differ between English and British English and introducing additional functional words. We make no changes to the hyperparameters used in the original method. 

\paragraph{Machine Translation (MT)} We used a round-trip machine translation method proposed by \citet{Keswani2016AuthorMT}. In this method, they translate the original text from English to German, German to French, and then French back to English. We enhanced their method by use of the new M2M translation model \cite{m2m100}, which does not rely on English as an intermediate language. 

\paragraph{JAMDEC (JD)} This method was proposed by \citet{fisher2024jamdec} and uses a small language model, GPT2-XL \cite{gpt2}, as the base model. For this method, they use a three stage approach where they extract the keywords of text (to guide generation to have the same content), overgenerate using diverse constrained beam search, and then filter based on grammar and content overlap. We used this model's default parameters, with a beam width of $10$, and only using the likelihood keyword extractors, which was recommended to be just as effective but take less time. More details of this methods' implementation can be found \cite{fisher2024jamdec}.

\paragraph{Paraphrasing} We used the paraphrasing model from \citet{jung2024impossible}. This model uses Google T5 \cite{t5} as the base and is finetuned on the dataset DIMPLE, which is a dataset of $4M$ high-quality pairs of paraphrases.  

\paragraph{Instruction LLMs} Lastly, we wanted to compare with LLMs of similar and bigger sizes. For these, we opted to use instruction tuned models which could easily follow instruction to rewrite the text. For each model, we used a temperature of $1.0$ and a top-p of $0.9$. \cref{tab:baseline_llm_instructions} shows the exact prompts used to generate the generations from each of the baseline LLMs.
\begin{table*}
\begin{adjustbox}{width=\textwidth}
\begin{tabular}{ ll }
\toprule
\textbf{Model} & \textbf{Instruction}\\
\midrule
Llama 2 &  "[INST] <<SYS>>\textbackslash n You are a helpful assistant.\textbackslash n \textbackslash n <</SYS>> \textbackslash nPlease rewrite the following: <paragraph>[/INST] Rewrite: " \\
Llama 3 &  "[INST] <<SYS>> \textbackslash n You are a helpful assistant.\textbackslash n \textbackslash n<</SYS>> \textbackslash nPlease rewrite the following: <paragraph>[/INST] Rewrite: "\\
Gemma & "You are a helpful assistant.\textbackslash n \textbackslash nPlease rewrite the following: <paragraph> Rewrite: "\\
Mistral & "<s>[INST] You are a helpful assistant.\textbackslash n \textbackslash nPlease rewrite the following: <paragraph> [/INST] Rewrite: " \\
\bottomrule
\end{tabular}
\end{adjustbox}
\caption{The instruction used for prompting the LLMs used as baselines. }
\label{tab:baseline_llm_instructions}
\end{table*} 
\subsubsection{\ourmethod}
\paragraph{Style and Weight Selection} We used the described automatic style and weight selection described in the paper and in \cref{appx:method_details}. We note that almost all values were less than $3$ standard deviations, with the majority between $0-2$.

\paragraph{Adapter Merging} We used three different ablation of our methods; sequential, adapter merging base, and adapter merging LoraHub+. For the sequential method, we averaged results over $n=3$ random shuffling of style axes orders. For the adapter merging base method we used the weight adapter found from mapping using the standard deviations. 

For the adapter merging LoraHub+, we build on the prior LoraHub method \cite{Huang2023LoraHubEC}. We used the weights selected using our mapping method as the initial values and then used a non-gradient based optimization \cite{liu2020versatile} over a new customized loss function. The loss function adds together the automatic evaluations from the author vector (described in \cref{appx:our_method_style_weight_selection}) for the specific style axes that are being considered for merging. Note, that since we are optimizing by finding the lowest loss, if the direction of the style axes is "higher" we take $1-value$ and if the direction is "lower" we just add the $value$. Lastly, we also add the grammar score into the loss to maintain good fluency. Then, a non-gradient based optimization method is use \cite{liu2020versatile}. Note, we use non-gradient based due to the large number of parameters of the model. We provide a comparison of the base weights chosen and the optimized LoraHub weights in \cref{tab:base_vs_lorahub_weights}. 

\paragraph{Hyperparameter Selection}
To tune the hyperparameters of \ourmethod, we use the validation split of \ourdataset.

\begin{table}[ht]
\centering
\footnotesize
\begin{tabular}{ c c }
\toprule
\textbf{\# of Std. Deviations} & \textbf{Adapter Weight}\\
\midrule
0-1 & 0.7 \\
1-2 & 0.9 \\
2-3 &  1.2\\
3+ & 1.5 \\
\bottomrule
\end{tabular}
\caption{This shows the static mapping used in the base adapter merging method. We use the number of standard of deviations an authors automatic style score is from the average style score of all authors in that domain. The static values were chosen base on past work \cite{Huang2023LoraHubEC}}
\label{tab:std_to_adapter_weight_mapping}
\end{table} 
\subsection{Evaluation Methodology and Other Details}\label{appx:exp_eval}
\paragraph{Obfuscation: Classifier} 
\label{obf_appendix_classifier}
We train classifiers over each of the four domains in \ourdataset to measure obfuscation during evaluation using their respective training and development sets. Specifically, for each of \ourdataset - $\{$ speech, novels, scholar, blog $\}$, we train a RoBERTa-Large classifier \cite{Liu2019RoBERTaAR} with a learning rate of 5e-5, batch size of 64, seed of 0, a max length of 256, and for 10 epochs. We set an early stopping threshold of 5, and choose the best checkpoint based on the best evaluation accuracy product (to ensure high accuracy for all classes). 

Overall, our final evaluation accuracy products for \ourdataset - $\{$speech, novels, scholar, blog $\}$ are 74.5\%, 85.6\%, 100\%, and 70.8\%, while the average overall accuracies are 90.6\%, 95.8\%, 100\%, and 93.3\%. Further training details including loss functions can be found in our repository. We train each of these models with a single NVIDIA A100 80 GB GPU for approximately 2 hours.

\label{appx:obf_classifier}
\paragraph{Content Preservation: Cosine Similarity} 
We compute neural text embeddings on the inputs and their obfuscations in Sentence Transformers \cite{Reimers2019SentenceBERTSE}. Next, we use the cosine similarity between the two vectors to gauge semantic similarity and get a approximation of content preservatinon. Note that though the cosine similarity can output values from -1 to 1, we find on all of our validation dataset (across all datasets and methods) \textbf{all} similarities between inputs and their obfuscations are non-negative, with a bound of 0 to 1. If the similarity 
metric were to, in a very rare case, have a negative value, we would set the value to 0 so that we could have a still meaningful overall product of metrics; however, we never observe this.

\paragraph{Grammar: CoLA} To ensure both fluency and grammaticality, we use TextAttack \cite{textattack}, a RoBERTa-large model \cite{Liu2019RoBERTaAR} fine-tuned on the Corpus of Linguistic Acceptability (CoLA; \citealp{Warstadt2018NeuralNA}) which includes 10,600 sentences with binary annotations for linguistic acceptability.

\section{Human Evaluation}\label{appx:appx:human_evals}
We omit the MT and Stylo methods from human evaluation as \citet{fisher2024jamdec} show that JamDec outperforms them in previous work for both automatic and human evaluation. We also omit 13B and 70B models for fair comparison. Finally, we report human evaluation for the best performing \ourmethod.

\begin{figure*}[t]
    \centering
    \includegraphics[trim={5cm 0 6cm 0},   width=0.55\textwidth]{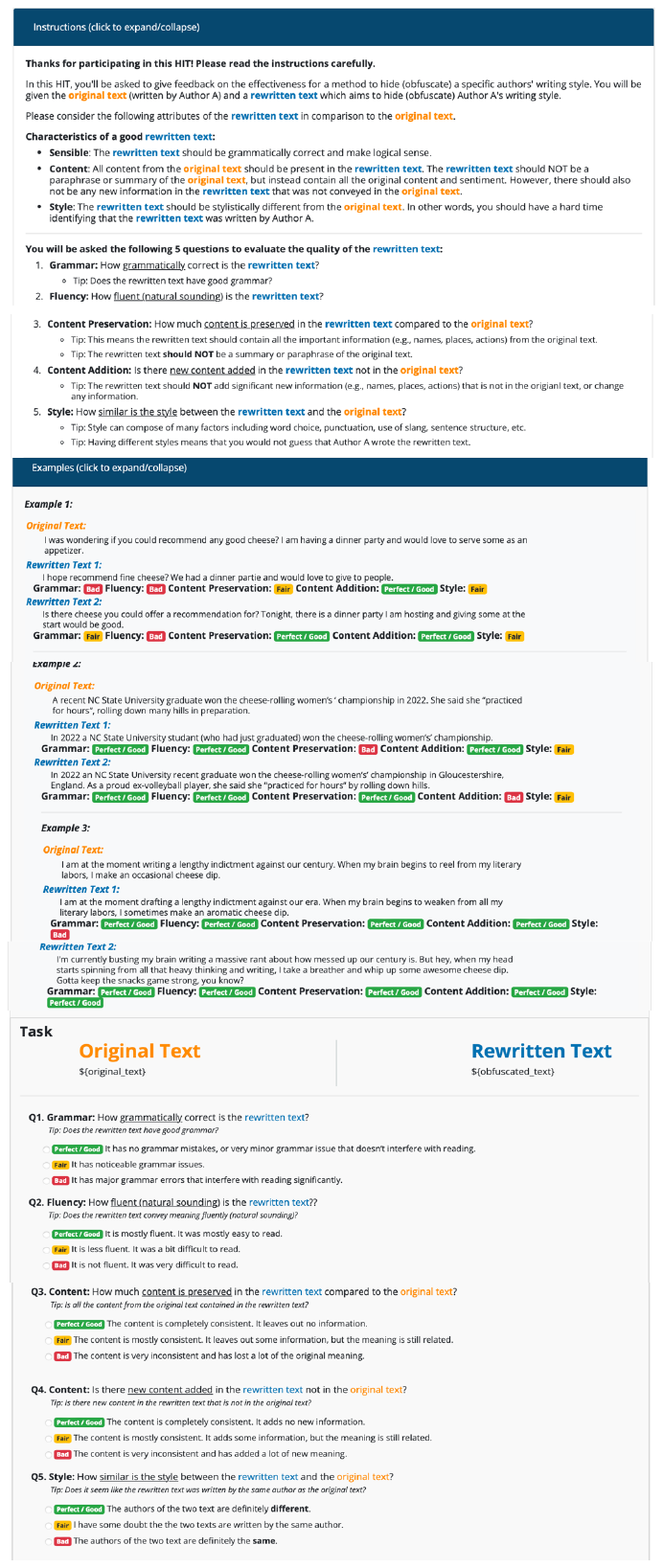}
    \caption{Instructions used in the human evaluation}
    \label{fig:human evaluation}
\end{figure*}

We used workers from Amazon Mechanical Turk (AMT) who voluntarily opt-in to the task to annotate $n=20$ text from each author. Each text was annotated by $n=3$ authors, who were paid at a rate of $15\$$/hour. Our annotators are from English-speaking countries. A screenshot of the interface is shown in \cref{fig:human evaluation}

Our agreement numbers for the five metrics we collect, grammaticality, fluency, content preservation, low content addition, and obfsucation are 99.8, 98.2, 95.3, 93.8, and 93.5\% respectively.

To compute the overall score, it would be unfair to take the product of our metrics (grammaticality, fluency, content preservation, low content addition, and obfuscation), since content preservation and fluency have two metrics representing their category, while obfuscation has only one; rewrites that simply output fluent, content-preserving texts would score higher than more balanced obfuscations which sacrifice small amounts of content preservation. Instead, we take a \emph{weighted} product. Our overall product is computed as:

{\footnotesize
\begin{align*}
\mbox{overall} = &\frac{\mbox{grammaticality} + \mbox{fluency}}{2} \times \\
& \frac{\mbox{content preservation} + \mbox{low content added}}{2} \times \\
&\text{obfuscation}
\end{align*}
}%
This product fairly takes the product of the three major categories (fluency, content preservation, and obfuscation), which aligns with our automatic metrics.

\section{Alternative Obfuscation Evaluation Metrics}
\label{altmetrics}

To verify the obfuscation effectiveness of \ourmethod, we run an alternative evaluation to measure drop rate using the method from Learning Universal Authorship Representations (LUAR; \citealp{Soto2021LearningUA}) and train models to learn authorship embeddings for each of the four domains in our AuthorMix dataset (speech, novels, scholar, and blog) using the training data. We use the default hyperparameters from the codebase (ie, training for 20 epochs, using sentence-transformers/paraphrase-distilroberta-base-v1 as the base model, etc).

Next, we create authorship embeddings for all authors by passing their validation data into the trained models with their respective domain where they are aggregated, resulting in a single embedding for each author. To perform authorship attribution and obtain predictions for a set of input data of size $N$ over some domain (such as speech), we first pass the input data through the trained model to extract individual embeddings before they are aggregated, resulting in $N$ input embeddings. For each of these input embeddings, we calculate the cosine similarity with each validation authorship embedding in the specified domain. The predicted authorship style is the one with the highest cosine similarity.

As with our RoBERTa models, we can compute the drop rate for the LUAR method. Recall that the drop rate is the drop in accuracy of the classifier evaluated on the original text and the obfuscated text, where accuracy represents how many of the text the classifier correctly identified the author. For each domain in AuthorMix, we obtain the initial classification accuracy with the LUAR method using the test set. Then, we calculate the LUAR drop rate for our StyleRemix methods and for the baselines.

In \cref{tab:auto_eval_baseline_luar}, we display \cref{tab:auto_eval_baseline} from the main obfuscation results, but now add two new rows which we compute: [NEW] Drop Rate w/ LUAR, the drop rate with LUAR, and [NEW] Overall (using Drop Rate w/ LUAR), the overall obfuscation quality, equivalent to the product of the grammar, content and Drop Rate w/ LUAR which we compute.

Across all datasets, the new authorship attribution results with LUAR aligns with our RoBERTa based results and reinforces the strong obfuscation results of StyleRemix over baselines. Specifically, across all datasets, StyleRemix has the highest LUAR drop rate on the speech, novels, and scholar datasets, and the second-highest LUAR drop rate on the blog dataset, beating much larger baselines like Llama-3-70b-Inst. This is the same as the previous obfuscation results and underlines the effectiveness of StyleRemix for obfuscation. Furthermore, the new overall metric with LUAR drop rate confirms the results from the original overall metric: over all datasets, StyleRemix generates the best overall obfuscations, beating all baselines.

Overall, our additional evaluation using LUAR authorship attribution confirm the previous result obtained with the RoBERTa classifiers and demonstrates the excellent anonymization capabilities of StyleRemix.

\begin{table*}[t!]
\begin{adjustbox}{width=\textwidth}
\begin{tabular}{lcccccccccccc}
\toprule
\textbf{Model}     & \multicolumn{2}{c}{\textbf{Llama-2-Chat}}            & \multicolumn{2}{c}{\textbf{Llama-3-Inst}}          & \multicolumn{1}{l}{\textbf{Gemma-Inst}} & \multicolumn{1}{l}{\textbf{Paraphrase}} & \multicolumn{1}{l}{\textbf{MT}} & \multicolumn{1}{l}{\textbf{Stylo}} & \textbf{JD} & \multicolumn{3}{c}{\ourmethod} \\
\cmidrule(lr){2-3} \cmidrule(lr){4-5} \cmidrule(lr){6-6} \cmidrule(lr){11-13}
\textbf{Size}      & \multicolumn{1}{c}{\textbf{7B}}     & \multicolumn{1}{c}{\textbf{13B}} & \multicolumn{1}{c}{\textbf{8B}}     & \multicolumn{1}{c}{\textbf{70B}} & \multicolumn{1}{c}{\textbf{7B}}    & \multicolumn{1}{l}{}                     & \multicolumn{1}{l}{}            & \multicolumn{1}{l}{}               &             & \multicolumn{1}{l}{\textbf{Seq.}} & \multicolumn{1}{l}{\textbf{AM}} & \multicolumn{1}{l}{\textbf{AM + LoraHub$^*$}} \\
         \rowcolor{gray!25}
        \multicolumn{13}{c}{\ourdataset -- \textbf{Speech}} \\
\textbf{Drop Rate} & 18.2   & 24.0 & 17.6&16.8    & 23.1  & 24.1 & 10.3 & 15.1  &   29.2   & 34.9& 41.2 & 31.4     \\
\textbf{[NEW] Drop Rate w/ LUAR} & 8.3& 7.4& 4.7& 6.8& 5.3& 3.2& 0.0& 7.7& 9.2  & 3.3& 23.9 & 12.2    \\
\textbf{Grammar}   & 67.8   & 67.1 &67.1 &70.2    & 67.8  & 71.2 & 54.9 & 37.8  &   56.7   & 61.7& 66.5 & 63.9     \\
\textbf{Content}   & 83.8   & 80.8 &80.8 &80.2    & 78.6  & 83.9 & 89.1 & 89.5  &    56.4 & 71.3& 77.3 & 73.9     \\
\textbf{Overall}   & 10.3   & 13.0 & 9.5&9.5    & 12.3  & 14.4 & 5.1&  5.1   &  9.4   & \underline{15.3}  & \textbf{21.2} & 14.8     \\ 
\textbf{[NEW] Overall (using Drop Rate w/ LUAR)}   & 4.7  & 4.0 & 2.5&3.8    & 2.8 & 1.5 & 0.0 &  2.6   &  2.9  & 1.5  & \textbf{12.3} & \underline{5.8}    \\ 
  \rowcolor{gray!25}
 \multicolumn{13}{c}{\ourdataset -- \textbf{Novels}} \\
\textbf{Drop Rate} & 12.2   & 13.7 &   9.2  & 11.3 & 13.3  & 10.8 & 7.0& 13.5  &    24.9  & 19.3& 28.6 & 35.6     \\
\textbf{[NEW] Drop Rate w/ LUAR} &3.6& 5.0& 5.7& 3.6& 6.4& 5.3& 2.2& 10.4& 16.5& 8.8& 17.9 & 31.7     \\
\textbf{Grammar}   & 71.8   & 73.8 &  73.1  & 75.4 & 70.0  & 68.3 & 46.3 & 36.8  &   61.2   & 72.6& 68.1 & 63.5     \\
\textbf{Content}   & 82.9   & 80.7 &   83.1 & 81.5 & 81.9  & 81.3 & 85.2 & 88.1  &    58.6  & 83.7& 76.1 & 72.9     \\
\textbf{Overall}   & 7.3    & 8.2&   5.6  & 6.9& 7.6   & 6.0  & 2.8& 4.4   &    8.9  & 11.8& \underline{14.8} & \textbf{16.5}  \\
\textbf{[NEW] Overall (using Drop Rate w/ LUAR)}   & 2.1    & 3.0&  3.5   & 2.2 & 3.7 & 2.9 & 0.9& 3.4  &   5.9  & 5.3 & \underline{9.3} & \textbf{14.7}
\\ \midrule
         \rowcolor{gray!25}
        \multicolumn{13}{c}{\ourdataset -- \textbf{Scholar}} \\
\textbf{Drop Rate} & 0.8    & 1.5& 1.6    & 2.5& 0.0   & 0.8  & 1.5& 4.6   &   6.1   & 1.8 & 9.2& 11.5     \\
\textbf{[NEW] Drop Rate w/ LUAR} & 6.1& 2.6& 5.2& 6.1& 6.9& 0.9& 0.0& 1.8& 5.2   & 10.4 & 10.4 & 13.0 \\
\textbf{Grammar}   & 64.3   & 64.9 & 64.1   & 66.6 & 65.3  & 69.1 & 54.5 & 31.0  &   62.3   & 65.8& 48.6 & 44.7     \\
\textbf{Content}   & 91.7   & 89.7 & 88.9   & 84.0 & 88.9  & 91.3 & 92.8 & 85.8  &   60.6  & 78.0& 75.3 & 68.8     \\
\textbf{Overall}   & 0.5    & 0.9& 0.9    & 1.4& 0.0   & 0.5  & 0.8& 1.2   &  2.3  &      0.9  &\underline{3.4}& \textbf{3.5}      \\
\textbf{[NEW] Overall (using Drop Rate w/ LUAR)}   & 3.6    & 1.5 & 3.0    & 3.4 & \underline{4.0}   &  0.6 & 0.0 & 0.5   &  2.0  &      \textbf{5.3} & 3.8 & \underline{4.0} \\
\midrule
         \rowcolor{gray!25}
        \multicolumn{13}{c}{\ourdataset -- \textbf{Blog}} \\
\textbf{Drop Rate} & 17.7 & 21.3 & 21.8  & 18.9 & 27.5 & 22.2  & 9.4 & 12.1  &   56.4   & 34.4& 41.0 & 42.0     \\
\textbf{[NEW] Drop Rate w/ LUAR} & 7.2& 9.4& 7.5& 5.9& 12.3& 14.2& 6.0& 9.1& 19.4  & 12.7 & 16.2 &14.2     \\
\textbf{Grammar}   & 68.4   & 69.1 & 71.3   & 74.0 & 69.0  & 69.8 & 41.9 & 29.1  &   60.6  & 66.7& 64.9 & 65.3     \\
\textbf{Content}   & 82.5   & 79.0 & 78.1   & 77.8 & 77.8  & 80.4 & 83.7 & 85.8  &   45.1   & 72.1& 73.7 & 74.2     \\
\textbf{Overall}   & 10.0   & 11.6 & 12.1   & 10.9 & 14.8  & 12.5 & 3.3& 3.0   &   15.4   & 16.5& \underline{19.6} & \textbf{20.4}    \\
\textbf{[NEW] Overall (using Drop Rate w/ LUAR)}   & 4.1  & 5.1 & 4.2   & 3.4 & 6.6  & 8.0 & 2.1 & 2.3   &   5.3   & 6.1 & \textbf{7.7} & \underline{6.9}   \\
\bottomrule
\end{tabular}
\end{adjustbox}
    \caption{Comparison of obfuscation methods measured by 1) mean drop rate, grammar, meaning similarity, and overall (the same metrics and results as in \cref{tab:auto_eval_baseline}) and 2) additionally with \textbf{an alternative, LUAR-based drop rate metric} and a new overall score computed with this drop rate. \textbf{Bold} and \underline{underline} denote the highest and the second-highest score respectively in each row.  All metrics displayed in the table are multiplied by 100 for easier viewing of significant figures.}
    \label{tab:auto_eval_baseline_luar}
\end{table*} 

\clearpage
\onecolumn
\section{Further Examples}
\begin{longtable*}{p{.15\textwidth}p{.85\textwidth}}
\toprule
         \rowcolor{gray!25}
        \multicolumn{2}{c}{\textbf{\ourdataset-Speech}} \\ \midrule
        \rowcolor{gray!15}
\textbf{Author}      & Trump                                                                                                                                                                               \\
\textbf{Original}    & Americans know that in a world where others seek conquest and domination, our nation must be strong in wealth, in might, and in spirit. That is why the United States vigorously defends the traditions and customs that have made us who we are.                       \\
\textbf{MT}          & The Americans know that in a world where the other is seeking conquest and domination, our people must be strong in wealth, power and spirit. That's why the United States defends the traditions and habits that have made us know who we are.                         \\
\textbf{Stylo}       & However, Americans know that in a world where others seek conquest and domination, our nation must be strong in wealth, in might, and in spirit. Lastly, That is why the United States vigorously depends the traditions and customs that have made us who we are.      \\
\textbf{JAMDEC} & Americans know that in a world where others seek conquest and domination, our nation must be strong in wealth, in might, and in spirit. That is why the United States vigorously defends the traditions and customs that have made us who we are. \\

\textbf{Paraphr}     & In a world where others strive for dominance and control, Americans recognize the importance of being strong in wealth, strength, and spirit. As a result, the United States vigorously protects the cultural and historical heritage that has contributed to our identity.                                                                                                                                                                                                                                                                                                                                           \\
\textbf{Llama2 (7B)} & As Americans, we recognize the importance of a strong and resilient nation in a world where power-hungry individuals and nations seek to conquer and dominate. To protect our identity and values, the United States fiercely guards the traditions and customs that define us as a nation. These beliefs are the bedrock upon which our great country was built, and they continue to serve as the foundation for our continued prosperity and influence in the world today.                                                                                                                                         \\
\textbf{Llama3 (8B)} & Americans understand that in a world where some nations strive for dominance, it is crucial that our own nation remains robust in its economic, military, and moral strength. This is why the United States actively preserves and upholds the values, customs, and traditions that have shaped our identity and character.                                                                                                                                                                                                                                                                                           \\
\textbf{Mistral V2}  & Americans recognize the importance of maintaining strength in wealth, military power, and cultural values in a global landscape where others may strive for conquest and dominance. Consequently, the United States is committed to safeguarding its heritage and customs that define its identity.                                                                                                                                                                                                                                                                                                                   \\
\textbf{Gemma (7B)}  & In a world where others strive for dominance, Americans believe their nation must maintain strength in wealth, military power, and morale. To uphold their heritage and identity, the United States stands firm in protecting its cherished traditions and customs.     \\
        \rowcolor{gray!5}
\textbf{\ourmethod}    & Americans hold the conviction that, in a universe where many nations pursue expansion and hegemony, our country must maintain its robustness in wealth, military prowess, and unwavering resolve. Hence, the United States zealously protects the very traditions and values that have shaped our distinct identity.                                                                                                                                                                                                                                                                                                  \\
        \rowcolor{gray!15}
\textbf{Author}      & Obama                                                                                                                                                                                       \\
\textbf{Original}    & Thank you. God bless you. And God bless this country we love.                                                                                                                               \\
\textbf{MT}          & Thanks for you. God blessed you. And God blesses this country that we love.                                                                                                                 \\
\textbf{Stylo}       & Thank you, god bless you; and and God bless this country we love.                                                                                                                           \\
\textbf{JAMDEC} &  And because we are on the cusp of a new era in human history, one that will be defined by our ability to boldly go where no one has gone before, to reach for new frontiers, to seek out new life and new civilizations, and to learn from the mistakes and successes of the past so that we might better prepare ourselves for the challenges of  And because, in the words of President Kennedy, we choose to go to the moon in this decade and do the other things, not because they were easy, but because they are hard, because that challenge is one that we are willing to accept, one we are unwilling to postpone, and one which we intend to win, and the others, too."\\
\textbf{Paraphr}     & Thank you, God bless you, and God bless this nation we cherish.                                                                                                                             \\
\textbf{Llama2 (7B)} & Thank you. May God bless you and our country.                                                                                                                                               \\
\textbf{Llama3 (8B)} & I would like to express my gratitude to you. May God grant you His blessings. May God also continue to bless our beloved nation.                                                            \\
\textbf{Mistral V2}  & I appreciate your kind words. May God's blessings be upon you. And may God continue to bless the country we cherish deeply.                                                                 \\
\textbf{Gemma (7B)}  & Thank you. May God bless you and this beloved nation.                                                                                                                                       \\
\rowcolor{gray!5}
\textbf{\ourmethod}    & Thank you. May God bless you, and may He continue to bless this nation that we hold dear.                                                                                                   \\
        \rowcolor{gray!15}
\textbf{Author}      & Bush                                                                                                                                                                                        \\
\textbf{Original}    & His tax -- his policies of tax and spend -- of expanding government rather than expanding opportunity -- are the policies of the past. We are on the path to the future -- and we're not turning back.                                                                  \\
\textbf{MT}          & His taxes ‚his tax and expenditure policy ‚the extension of the government rather than the extension of possibilities ‚are the policies of the past. We are on the way to the future ‚and we will not return.                                                           \\
\textbf{Stylo}       & His tax-- his policies in tax and spend-- in expanding government rather than expanding opportunity-- are the policies in the past. We are on the path to the future-- and we 're not turning back.                                                                     \\
\textbf{JAMDEC} & His tax -- his policies of tax and spend -- of expanding government rather than expanding opportunity -- are the policies of the past. We are on the path to the future -- and we're not turning back.  \\
\textbf{Paraphr}     & The current tax policies, which prioritize government expansion over promoting economic growth, are a reflection of the past. However, we are on the path to the future and we will not retreat.                                                                        \\
\textbf{Llama2 (7B)} & His tax and spend policies, which focused on expanding the government rather than creating opportunities for growth, are a relic of the past. We are moving forward with a renewed commitment to innovation and progress, leaving behind the stagnant policies of the past.                                                                                                                                                                                                                                                                                                                                           \\
\textbf{Llama3 (8B)} & His tax philosophy of spend and tax, which prioritizes government expansion over creating opportunities, is a relic of the past. We're now on a trajectory towards a brighter future and we won't be reversing course.                                                  \\
\textbf{Mistral V2}  & His outdated tax and spending policies, which focus more on expanding government rather than creating opportunities, are no longer relevant. We are moving forward towards the future and will not revert to these old ways.                                            \\
\textbf{Gemma (7B)}  & His tax and spending policies, which aim to expand government rather than foster opportunity, reflect outdated ideologies. We are committed to embracing a forward-looking vision that prioritizes progress and innovation.                                             \\
\rowcolor{gray!5}
\textbf{\ourmethod}    & He's all about taxing up a storm and spending it all, plus, he's all about growing government big instead of creating more chances for everyone. That's really old-school thinking. We're all about moving forward and not looking back.                                \\
\midrule
         \rowcolor{gray!25}
        \multicolumn{2}{c}{\textbf{\ourdataset-Novels}} \\ \midrule
        \rowcolor{gray!15}
\textbf{Author}      & Hemingway                                                                                                                                                                                   \\
\textbf{Original}    & "Don't worry about money," Mike said. ‚"You can pay for the car, Jake, and I'll send you my share."                                                                                         \\
\textbf{MT}          & "They don't worry about money,"Mike said. ‚"You can pay for the car, Jake, and I'll tell you my part."                                                                                      \\
\textbf{Stylo}       & " Don ' t worry about money, "Mike said, and ‚" You can pay for the wagon, Jake, and I ' ll send you my party, and "                                                                        \\

\textbf{JAMDEC} &   "We can pay the family and move on with our lives, and we won't have to worry about where we're going to live or who's going to take care of us when we're gone.\\

\textbf{Paraphr}     & Mike advised Jake not to worry about finances. He suggested that Jake could pay for the car himself, and he would then send him his share of the cost.                                      \\
\textbf{Llama2 (7B)} & "Don't worry about money," Mike said. "I'll cover my share of the cost, and you can pay for Jake's."                                                                                        \\
\textbf{Llama3 (8B)} & Mike reassured Jake, "Don't stress about the cost, Jake. You take care of buying the car, and I'll cover my portion of the expense."                                                        \\
\textbf{Mistral V2}  & Mike assured Jake, "You don't need to be concerned about the cost of the car. I'll cover my portion and send it to you later."                                                              \\
\textbf{Gemma (7B)}  & "Don't worry about the cost,"Mike said. ‚"I'll cover the car, Jake, and I'll transfer my share to you later."                                                                               \\
\rowcolor{gray!5}
\textbf{\ourmethod}    & Mike addressed the financial concerns with a seemingly casual remark. "Don't stress over finances," he assured, "Jake, you handle the car expenses, and I'll cover my portion." This approach, seemingly nonchalant about the financial management, was delivered with a casual air.                                                                                                                                                                                                                                                                                                                                  \\
        \rowcolor{gray!15}
\textbf{Author}      & Fitzgerald                                                                                                                                                                                  \\
\textbf{Original}    & She liked him immediately‚ reddish-brown mustache under watery blue eyes that had something in them that these other eyes lacked, some quality of appreciation. They exchanged stray sentences through dinner, and she made up her mind to see him again.               \\
\textbf{MT}          & She loved her immediately ‚a red-brown moustache under the blue eyes of water that had something in them, which was missing in those other eyes, a kind of appreciation. They exchanged stray for dinner, and she awakened her mind to see him again.                   \\
\textbf{Stylo}       & She liked her immediately‚ a reddish-brown mustache, under watery blue eyes that had something of them that these other eyes lacked, some quality of appreciation. They exchanged stray sentences through supper, and she made up her mind to see it again.             \\
\textbf{JAMDEC} & \\
\textbf{Paraphr}     & She immediately fell in love with him, noticing his distinctive reddish-brown mustache beneath his watery blue eyes, which had a unique quality that set him apart from other eyes. They exchanged brief sentences during dinner, and she made up her mind to see him again.                                                                                                                                                                                                                                                                                                                                          \\
\textbf{Llama2 (7B)} & She was immediately taken by him, noticing the reddish-brown mustache framing his watery blue eyes. There was something unique about those eyes, a quality of thoughtfulness that set them apart from the other eyes she had seen. They exchanged brief words throughout dinner, and she found herself making plans to see him again.                                                                                                                                                                                                                                                                                 \\
\textbf{Llama3 (8B)} & As soon as she met him, she was drawn to him - the distinctive reddish-brown mustache and watery blue eyes that seemed to hold a certain depth of understanding. Throughout dinner, they exchanged brief, scattered phrases, and she found herself resolved to reconnect with him again in the future.                                                                                                                                                                                                                                                                                                                \\
\textbf{Mistral V2}  & From the first moment, she was drawn to him. His reddish-brown mustache framed watery blue eyes that held a unique quality of appreciation. During dinner, they engaged in brief conversations, and she found herself looking forward to another encounter with him.    \\
\textbf{Gemma (7B)}  & She instantly admired his reddish-brown mustache framing his watery blue eyes, which held a depth that set them apart from the others. They exchanged fleeting words throughout dinner, and she resolved to meet him again.                                             \\
\rowcolor{gray!5}
\textbf{\ourmethod}    & She felt an instant connection with him. His reddish-brown mustache framed his watery blue eyes, which seemed to hold a special kind of appreciation, different from others she had seen. During dinner, they shared a few casual conversations, and she decided she wanted to meet him again.                                                                                                                                                                                                                                                                                                                        \\
        \rowcolor{gray!15}
\textbf{Author}      & Woolf                                                                                                                                                                                       \\
\textbf{Original}    & "But I must look so queer!"she cried, running over to the glass and looking first this side then that. Then she snatched it off again, for there was a tap at the door. Could it be Sir William Bradshaw? Had he sent already?                                          \\
\textbf{MT}          & "But I need to look so beautiful!"she cried, blown to the glass and looked first on that side, then on that. Then she pushed him, because there was a ball at the door. Can he be Sir William Bradshaw? Has he already sent?                                            \\
\textbf{Stylo}       & " But I must listen so queer; and "she cried, running over to the drink and looking first this side then that, then she snatched it off once again, for there was a tap at the door; could it be Sir William Bradshaw, had he sent already?                             \\
\textbf{JAMDEC} &  she cried, looking at herself in the mirror. "I don't know what's the matter with me.  "What is the matter with me?" she asked, looking at herself in the glass.   She looked again, and there he was, sitting in a chair by the table.\\

\textbf{Paraphr}     & She exclaimed, "I must appear unusual!" as she ran towards the glass and carefully examined it, first on one side and then the other. Then she quickly snatched it off again, as there was a tap at the door. Could it be Sir William Bradshaw? Had he already sent something?                                                                                                                                                                                                                                                                                                                                        \\
\textbf{Llama2 (7B)} & "But I must look so strange!" she exclaimed, rushing over to the mirror and examining herself from different angles. She then quickly removed the mirror from the wall, as she heard a knock at the door. "Could it be Sir William Bradshaw? Has he arrived already?" she wondered aloud.                                                                                                                                                                                                                                                                                                                             \\
\textbf{Llama3 (8B)} & But I must look so peculiar! she exclaimed, hurrying to the mirror and scrutinizing herself from every angle. She then hastily removed the mirror, her attention distracted by a knock at the door. Could it be Sir William Bradshaw? Had he already arrived, sent by her aunt?                                                                                                                                                                                                                                                                                                                                       \\
\textbf{Mistral V2}  & "Oh dear, I must look quite strange!" she exclaimed, hurrying towards the mirror. She glanced at her reflection from various angles before hurriedly removing the mirror from the wall. A knock at the door interrupted her, and she wondered if it could be Sir William Bradshaw. Had he arrived already?                                                                                                                                                                                                                                                                                                            \\
\textbf{Gemma (7B)}  & In a sudden burst of anxiety, she exclaimed, "Oh, I must look so strange!" and rushed over to the glass. She peered cautiously through, casting her gaze first one way and then the other. Moments later, she snatched the glass away, for a tap at the door startled her. The unfamiliar sound made her wonder if it could be Sir William Bradshaw, and if he had already arrived.                                                                                                                                                                                                                                   \\
\rowcolor{gray!5}
\textbf{\ourmethod}    & She suddenly felt awkward when she noticed herself in the mirror, turning this way and that to check her appearance. Just as she was about to examine herself further, someone knocked on the door. She wondered if it might be Sir William Bradshaw, wondering if he had already arrived.                                                                                                                                                                                                                                                                                                                            \\
\midrule
         \rowcolor{gray!25}
        \multicolumn{2}{c}{\textbf{\ourdataset-Scholar}} \\ \midrule
        \rowcolor{gray!15}
\textbf{Author}      & Scholar-h                                                                                                                                                                                   \\
\textbf{Original}    & So what are the takeaways from the Nepal case study?                                                                                                                                        \\
\textbf{MT}          & So what are the Takeaways of the Nepal case study?                                                                                                                                          \\

\textbf{Stylo}       & So what are the takeaways from the Nepal case study?                                                                                                                                        \\
\textbf{JAMDEC} &  First of all, it is important to note that this is not the first time that the Nepalese government has had.\\
\textbf{Paraphr}     & The Nepal case study has provided valuable insights.                                                                                                                                        \\
\textbf{Llama2 (7B)} & What are the key takeaways from the Nepal case study?                                                                                                                                       \\
\textbf{Llama3 (8B)} & Here are the key lessons learned from the Nepal case study: Please let me know if this meets your expectations or if you would like me to make any changes!                                 \\
\textbf{Mistral V2}  & What key insights can be gained from the Nepal case study?                                                                                                                                  \\
\textbf{Gemma (7B)}  & What are the key lessons learned from the Nepal case study?                                                                                                                                 \\
\rowcolor{gray!5}
\textbf{\ourmethod}    & Imagine yourself in the heart of a Nepal case study‚Äîwhat exactly are the lessons you're likely to glean? The narrative unfolds there, rich with nuances and insights, each experience potentially etched into the global lexicon.                                     \\
        \rowcolor{gray!15}
\textbf{Author}      & Scholar-pp                                                                                                                                                                                  \\
\textbf{Original}    & Even if you can't eliminate identity theft, you can get instant fraud notifications any time your identity is used without your knowledge. Prevent identity theft with MyID.                \\
\textbf{MT}          & Even if you cannot eliminate identity theft, you can receive instant fraud messages at any time if your identity is used without your knowledge. Avoid identity theft with MyID.            \\
\textbf{Stylo}       & Conversely, Even if you can ' t eliminate identity theft, you can obtain instant fraud modifications any time your identity is used without your knowledge; prevent identity theft with MyID.                                                                           \\
\textbf{JAMDEC} &  If you have questions about your identity, or if you think someone is trying to use it without your permission, click here to get instant alerts from the Identity Theft Resource Center (ITRC) and the Federal Trade Commission (FTC) about identity theft and how to protect.  Identity theft is the theft of your identity by someone who uses your identity without your permission, such as by using your Social Security number to open a bank account or apply for a credit card, or by opening up a new bank account in your name and using it to receive.\\

\textbf{Paraphr}     & Despite the fact that it may not be possible to completely prevent identity theft, you can receive instant notifications of any fraudulent activity involving your identity without your knowledge. This helps to protect your identity and prevent any potential threats.                                                                                                                                                                                                                                                                                                                                            \\
\textbf{Llama2 (7B)} & Although identity theft cannot be completely eliminated, MyID offers instant fraud notifications whenever your identity is used without your knowledge. Protect yourself from identity theft with MyID.                                                                 \\
\textbf{Llama3 (8B)} & While it's impossible to completely eliminate the risk of identity theft, you can receive immediate alerts whenever your identity is used without your consent. Stay protected with MyID, your trusted identity monitoring solution.                                    \\
\textbf{Mistral V2}  & MyID allows you to receive immediate notifications if your identity is used without your consent, providing an extra layer of protection against identity theft. Although it's impossible to completely prevent identity theft, having this alert system can help you take swift action to minimize potential damage.                                                                                                                                                                                                                                                                                                 \\
\textbf{Gemma (7B)}  & Stay vigilant against identity theft with MyID. Receive instant fraud notifications whenever your identity is used without your knowledge, even if you can't completely eliminate the risk. \\
\rowcolor{gray!5}
\textbf{\ourmethod}    & Amidst the complexities of digital security, the safeguarding of your identity might be challenged beyond your control, yet, it is assured that you will be alerted to instances of suspected fraud almost instantaneously whenever your identity is invoked without your awareness. This protection is diligently offered through the mechanism of MyID, a strategic layer of defense, designed to help you actively prevent the encroachment of identity theft.                                                                                                                                                     \\
        \rowcolor{gray!15}
\textbf{Author}      & Scholar-qq                                                                                                                                                                                  \\
\textbf{Original}    & They further stated that an independent Angola under the MPLA would be dominated by urban mesticos and assimilados, effectively transferring power from one class of colonial masters to another.                                                                       \\
\textbf{MT}          & They continued to explain that an independent Angola under the MPLA is dominated by urban mesticos and assimilados, which effectively transfer power from one class of colonial masters to another.                                                                     \\
\textbf{Stylo}       & They further stated that an independent Angola under the MPLA would be dominated by urban mesticos and assimilados, effectively transferring power from one class of colonial masters to another.                                                                       \\
\textbf{JAMDEC} &  This was further stated in an article entitled "What would happen if Angola's urban elite became mestisos?" which was published in La Repblica, the official mouthpiece of the Communist Party of Angola (MPLA).\\

\textbf{Paraphr}     & According to the text, an independent Angola under the MPLA would be primarily governed by urban elites and assimilationists, effectively transferring power from one class of colonial leaders to another.                                                             \\
\textbf{Llama2 (7B)} & They further emphasized that an independent Angola under the MPLA would be dominated by urban mixed-race and assimilated individuals, effectively transferring power from one group of colonial oppressors to another.                                                  \\
\textbf{Llama3 (8B)} & According to them, if Angola became an independent country under the MPLA's leadership, it would likely be controlled by urban mesticos and assimilados, essentially replacing one group of colonial rulers with another.                                               \\
\textbf{Mistral V2}  & The MPLA's proposed independent Angola was predicted to be controlled by urban mesticos and assimilados, implying a shift in power from one colonial ruling class to another.               \\
\textbf{Gemma (7B)}  & According to their statement, an independent Angola under the MPLA would likely be controlled by urban dwellers who had mixed ancestry and those who had assimilated into Angolan society, effectively replacing one class of colonial masters with another.            \\
\rowcolor{gray!5}
\textbf{\ourmethod}    & They explained that if Angola became independent, led by the MPLA, it might end up being mostly ruled by people who were born in cities and considered either mesticos or assimilados. This would mean that the power might just shift from one group of people who were once in charge, to another group who were also outsiders, but from a different background.                                                                                                                                                                                                                                                   \\
\midrule
         \rowcolor{gray!25}
        \multicolumn{2}{c}{\textbf{\ourdataset-Blog}} \\ \midrule
        \rowcolor{gray!15}
\textbf{Author}      & Blog-1                                                                                                                                                                                      \\
\textbf{Original}    & Costco is a phenomenon you have to visit to believe.  This is impulse buying taken to a new level.                                                                                          \\
\textbf{MT}          & Costco is a phenomenon that you need to visit to believe. This is an impulse for the purchase to a new level.                                                                               \\
\textbf{Stylo}       & osteo 's a phenomenon you have to visit to believe, that is impulse buying adopted to a new level.                                                                                          \\
\textbf{JAMDEC} & Costco is a phenomenon you have to visit to believe.  This is impulse buying taken to a new level. \\
\textbf{Paraphr}     & Costco is a unique and remarkable experience that requires a visit to fully appreciate. It's like a shopping experience that takes the concept of impulse buying to a whole new level.      \\
\textbf{Llama2 (7B)} & Costco is a truly unique and unforgettable experience that will leave you amazed and surprised at every turn. Known for its unparalleled impulse buying opportunities, this retail giant has taken the concept of spontaneous shopping to a whole new level. From its vast array of products to its unbeatable prices, a visit to Costco is an event that should not be missed. Whether you're a seasoned shopper or just looking for a fun day out, this incredible store is sure to delight and surprise you in ways you never thought possible. So come and experience it for yourself ‚you won't be disappointed! \\
\textbf{Llama3 (8B)} & "Costco is an extraordinary experience that must be seen to be fully appreciated. It's a retailer that takes impulse buying to a whole new level, making it easy to get caught up in the excitement of discovery and purchase."                                         \\
\textbf{Mistral V2}  & Visiting Costco is an experience unlike any other, showcasing impulse buying at its peak.                                                                                                   \\
\textbf{Gemma (7B)}  & Costco's sprawling warehouses and endless aisles entice customers with an unparalleled shopping experience, urging them to explore the realm of bulk buying and spontaneous purchases.      \\
\rowcolor{gray!5}
\textbf{\ourmethod}    & Costco is a place you really need to see to understand. It's a spot where you might just end up buying more than you planned.                                                               \\
        \rowcolor{gray!15}
\textbf{Author}      & Blog-2                                                                                                                                                                                      \\
\textbf{Original}    & If not, then not.  "How long ago since you ate anything proper, hmm?"                                                                                                                       \\
\textbf{MT}          & If not, then not. How long have you spent since you've eaten something good?                                                                                                                \\
\textbf{Stylo}       & If not, then not.`` How a lengthy time ago since you ate anything proper, him?''                                                                                                            \\
\textbf{JAMDEC} &  And if not, well, then, I guess I'll just have to go back to the drawing board and figure out what to.  I don't know what else to say, other than that I'm really happy that I got to be a part of it and.\\
\textbf{Paraphr}     & "How long has it been since you last consumed a nutritious meal, hmm?"                                                                                                                      \\
\textbf{Llama2 (7B)} & If not, then not. How long has it been since you've had a proper meal?                                                                                                                      \\
\textbf{Llama3 (8B)} & It seems you haven't eaten a substantial meal recently, have you?                                                                                                                           \\
\textbf{Mistral V2}  & If you haven't eaten a proper meal for some time now, hmm?                                                                                                                                  \\
\textbf{Gemma (7B)}  & If not, then not. How long ago was it since you ate something proper, hmmm?                                                                                                                 \\
\rowcolor{gray!5}
\textbf{\ourmethod}    & If not, then indeed, not. "How long has it been since you last consumed a proper meal?"                                                                                                     \\
        \rowcolor{gray!15}
\textbf{Author}      & Blog-3                                                                                                                                                                                      \\
\textbf{Original}    & Quick lunch.....I wanted Jasper's today though.  Their food is startin' to really be good......mmmmm perogies and quesadillas....and pizza....mmmmmmmmmmmm.....................foooood...  Bio class was pretty lame.                                                   \\
\textbf{MT}          & Breakfast, but I wanted Jasper's today. Your diet begins to be really good...mmmm perogies and quesadillas....and pizza....mmmmmm.............foooood... the biological class was pretty lame.                                                                          \\
\textbf{Stylo}       & Quick lunch ..... myself wanted paper 's today though; and their food is starting ' to really be good ...... mamma perogies also quesadillas .... and penza .... mmmmmmmmmmmm ..................... foooood... mio class was pretty lame.                               \\
\textbf{JAMDEC} &  I don't know if I'll be able to get it, but I wanted to make sure I got it in time for lunch, so I made a quick lunch out of some of the things I had left over from the last time I made lunch, and a few things I found in the pantry that I thought would be good for lunch that I didn't have time to make lunch with, and then.  I was going to be honest and say that I didn't really care for the food here, but it was Jasper's so I had to give it a try! I ordered the chicken and waffles and it was pretty good! The waffles were crispy on the outside and soft on the inside, and the chicken was cooked perfectly!! I would definitely come back here if I was in the area again!\\

\textbf{Paraphr}     & The author wanted to have a quick lunch at Jasper's today, as their food is starting to be quite good. They are particularly fond of their perogies and quesadillas, as well as their pizza. The author found the bio class to be quite uninteresting.                  \\
\textbf{Llama2 (7B)} & "Ah, a quick lunch... I was actually hoping to visit Jasper's today, their food has really been impressing me lately. I can't get enough of their perogies, quesadillas, and pizza... mm mm mm... delicious! Unfortunately, bio class was quite dull today. "       \\
\textbf{Llama3 (8B)} & I was craving Jasper's for lunch today, but unfortunately, I didn't get a chance to go. Their food has been getting better and better, and I've been eyeing their perogies, quesadillas, and pizza. Mmm, just thinking about it is making me hungry! Too bad I missed out. On a different note, bio class was pretty underwhelming today.                                                                                                                                                                                                                                                                             \\
\textbf{Mistral V2}  & I was considering having a quick lunch, but I had a craving for Jasper's today. Their food has significantly improved recently, and I find myself longing for their perogies, quesadillas, and pizza - oh, the deliciousness! However, my biology class left me feeling less than satisfied.                                                                                                                                                                                                                                                                                                                          \\
\textbf{Gemma (7B)}  & My quick lunch today was at Jasper's. Their food is seriously good! I highly recommend their perogies, quesadillas, and pizza. Bio class was unfortunately pretty boring.                   \\
\rowcolor{gray!5}
\textbf{\ourmethod}    & Today's lunch session was brief and I had a particular desire to visit Jasper's. Their culinary offerings are notably improving, notably including their distinctive perogies, quesadillas, and pizza, which are truly exceptional. The experience was quite underwhelming during my bio class.                                                                                                                                                                                                                                                                                                                       \\
        \rowcolor{gray!15}
\textbf{Author}      & Blog-4                                                                                                                                                                                      \\
\textbf{Original}    & Nowadays, this phase I have been going through is making me face all these quite brutally.  Because I choose it to be so.  I have never been a woman who takes things swiftly (which is, yet another choice).                                                           \\
\textbf{MT}          & Today, this phase I've done faces me with all this quite brutal. Because I chose to be like that. I've never been a woman who takes things quickly (i.e. another choice).                   \\
\textbf{Stylo}       & Afterwards, Nowadays this: step I have been going through is making me face all these quite brutally; and because I choose it to be so, i have never been a wife which takes things swiftly ( which is yet another choice).                                             \\
\textbf{JAMDEC} &  I don't know what to do about it, but I'm going to try my best to get through this phase as best as I can, so that I can get on with my life.  I don't know if I'll ever be able to look at myself in the mirror and be proud of what I look like, but at least I know that I'm not a.  I choose to live my life as a woman, even though I know it's not a choice I'm allowed to make, it's a choice my body makes for me, and.\\

\textbf{Paraphr}     & The current phase I am experiencing is causing me to confront these challenges with a brutal intensity, as I have never been a person who takes things lightly (which is another choice).   \\
\textbf{Llama2 (7B)} & In recent times, I have been grappling with a phase that is proving particularly challenging. Rather than navigating it with ease, I find myself facing these challenges head-on with a sense of brutal directness. This is a choice I have made, one that I have always been hesitant to embrace. As someone who prefers to take things slowly and carefully, I have never been one to rush into situations without fully considering the consequences. This cautious approach has led to a certain degree of struggle, but it is a choice that I stand by nonetheless.                                              \\
\textbf{Llama3 (8B)} & Lately, I've been confronting some harsh realities, and I have to admit that it's a deliberate choice I've made. I've never been one to rush through life's challenges, and that's a choice I've always made for myself.                                                \\
\textbf{Mistral V2}  & Currently, I'm experiencing this period in life quite intensely. I've decided to face its challenges head-on, although I've never been one to act quickly in the past.                      \\
\textbf{Gemma (7B)}  & Currently, I am navigating through a phase that has forced me to confront various challenges with blunt honesty. This stems from my conscious decision to approach things at a deliberate pace, a characteristic that aligns with my intrinsic personality.             \\
\rowcolor{gray!5}
\textbf{\ourmethod}    & These days, I'm really laying it all out there, hitting the tough stuff head-on, because that's just how I roll. It's definitely not my style to breeze through life; I've always been more of the slow-and-thorough type‚Äîanother deliberate choice of mine.          \\
        \rowcolor{gray!15}
\textbf{Author}      & Blog-5                                                                                                                                                                                      \\
\textbf{Original}    & Anyway, it looks great.  I painted the same color, but in flat instead of eggshell, and it's a big improvement.  I love the way flat paint looks.                                           \\
\textbf{MT}          & However, it seems great. I made the same color, but in flat instead of garlic, and it's a great improvement. I like how Flat Paint looks like.                                              \\
\textbf{Stylo}       & Anyway, it looks awesome, and i painted the same color colour, but in apartment instead in eggshell, and it ' s a big improved; and i love the way apartment flat paint looks.              \\
\textbf{JAMDEC} &  It looks like it's going to be a really good game.  I can't wait to see what you come up with for the rest of the series, and I'm really looking forward to seeing what you do with the other characters in the series!  Im going to try to get a few more of these in the.\\

\textbf{Paraphr}     & The paint job turned out great. I used the same color, but in a flat finish instead of eggshell, and it's a significant improvement. I love the way flat paint looks.                       \\
\textbf{Llama2 (7B)} & "Ah, I see! Painting the same color but in a flat finish instead of eggshell has made a significant difference. I completely agree, flat paint can give a room a sleek and modern look that's simply stunning. It's great to hear that you're enjoying the new finish!" \\
\textbf{Llama3 (8B)} & To be honest, I'm really pleased with how it turned out. I opted to use the same colour, but in a flat finish instead of eggshell, and it's made a significant difference. I'm a big fan of the flat paint look, and I think it's a real improvement.                   \\
\textbf{Mistral V2}  & The new look is excellent. I opted for flat paint instead of eggshell for the same color, and I'm thrilled with the outcome. Flat paint really enhances the appearance.                     \\
\textbf{Gemma (7B)}  & In any case, it looks fantastic! I opted for flat paint instead of eggshell in the same color, and the difference is remarkable. I adore the seamless and sophisticated look of flat paint. \\
\rowcolor{gray!5}
\textbf{\ourmethod}    & The appearance has been significantly improved. The same color was chosen, but the application was altered from eggshell to flat, which has been found to significantly enhance the aesthetic. There is a preference for the appearance of flat paint.      \\
\bottomrule
\caption*{More qualitative examples of different methods.}
\label{tab:qual_examples}
\end{longtable*} 
 \end{document}